\documentclass[10pt,twocolumn,letterpaper]{article}

\usepackage[pagenumbers]{cvpr} %

\usepackage{graphicx}
\usepackage{amsmath}
\usepackage{amssymb}
\usepackage{booktabs}
\usepackage[pagebackref,breaklinks,colorlinks]{hyperref}
\usepackage{multirow}

\usepackage{color,xcolor}
\usepackage{epsfig}
\usepackage{graphicx}
\usepackage{times}

\usepackage{comment}

\usepackage{pifont}

\usepackage{microtype}
\usepackage[font=small]{caption}
\usepackage{arydshln}
\usepackage{tabularx}
\usepackage{adjustbox}
\usepackage{array}
\usepackage{booktabs}
\usepackage{colortbl}
\usepackage{float,wrapfig}
\usepackage{hhline}
\usepackage{multirow}
\usepackage{subcaption} %
\usepackage[percent]{overpic}

\usepackage{stmaryrd}
\usepackage{breqn}

\usepackage{amsmath,amsfonts,amssymb}
\usepackage{duckuments}
\usepackage{amsmath}

\usepackage{bm}
\usepackage{nicefrac}
\usepackage{microtype}
\usepackage{dsfont}
\usepackage{changepage}
\usepackage{extramarks}
\usepackage{fancyhdr}
\usepackage{setspace}
\usepackage{soul}
\usepackage{xspace}
\usepackage{hhline}
\usepackage{algorithmicx}
\usepackage{algorithm}
\usepackage{algpseudocode}
\usepackage{pifont}
\usepackage{amssymb}

\usepackage{makecell}

\usepackage[pagebackref,breaklinks,colorlinks]{hyperref}

\usepackage{enumitem}
\usepackage[title]{appendix}

\def\x{{\mathbf{x}}}
\def\v{{\mathbf{v}}}

\def\c{{\mathbf{c}}}

\def\d{{\mathbf{d}}}
\def\f{{\mathbf{f}}}

\newcommand{\smallsim}{\smallsym{\mathrel}{\sim}\hspace{-.5mm}}

\makeatletter
\newcommand{\smallsym}[2]{#1{\mathpalette\make@small@sym{#2}}}
\newcommand{\make@small@sym}[2]{%
  \vcenter{\hbox{$\m@th\downgrade@style#1#2$}}%
}
\newcommand{\downgrade@style}[1]{%
  \ifx#1\displaystyle\scriptstyle\else
    \ifx#1\textstyle\scriptstyle\else
      \scriptscriptstyle
  \fi\fi
}
\makeatother

\newcommand*{\menlo}{\fontfamily{lmtt}\fontsize{9}{9}\selectfont }

\newcommand{\reffig}[1]{Figure~\ref{fig:#1}}
\newcommand{\refsec}[1]{Section~\ref{sec:#1}}
\newcommand{\refapp}[1]{Appendix~\ref{sec:#1}}
\newcommand{\reftbl}[1]{Table~\ref{tbl:#1}}

\newcommand{\refeq}[1]{Eqn.~\ref{eq:#1}}

\newcommand{\lblfig}[1]{\label{fig:#1}}
\newcommand{\lblsec}[1]{\label{sec:#1}}
\newcommand{\lbleq}[1]{\label{eq:#1}}

\newcommand{\ignorethis}[1]{}

\newcommand{\myparagraph}[1]{\vspace{1pt} \noindent \textbf{#1} \ }

\def\1{\bm{1}}

\DeclareMathOperator*{\argmin}{arg\,min}

\newcolumntype{L}[1]{>{\raggedright\let\newline\\\arraybackslash\hspace{0pt}}m{#1}}
\newcolumntype{C}[1]{>{\centering\let\newline\\\arraybackslash\hspace{0pt}}m{#1}}
\newcolumntype{R}[1]{>{\raggedleft\let\newline\\\arraybackslash\hspace{0pt}}m{#1}}

\newcommand{\ignore}[1]{}

\renewcommand*{\thefootnote}{\arabic{footnote}}

\makeatletter
\DeclareRobustCommand\onedot{\futurelet\@let@token\@onedot}
\def\@onedot{\ifx\@let@token.\else.\null\fi\xspace}

\def\etal{\emph{et al}\onedot}
\makeatother

\usepackage[capitalize]{cleveref}
\crefname{section}{Sec.}{Secs.}
\Crefname{section}{Section}{Sections}
\Crefname{table}{Table}{Tables}
\crefname{table}{Tab.}{Tabs.}

\def\confName{CVPR}
\def\confYear{2023}

\begin{document}

\title{Multi-Concept Customization of Text-to-Image Diffusion
}

\author{Nupur Kumari\textsuperscript{1}
\qquad
Bingliang Zhang\textsuperscript{2}
\qquad
Richard Zhang\textsuperscript{3}
\qquad
Eli Shechtman\textsuperscript{3}
\qquad
Jun-Yan Zhu\textsuperscript{1}\\
\textsuperscript{1}Carnegie Mellon University
\qquad
\textsuperscript{2}Tsinghua University
\qquad
\textsuperscript{3}Adobe Research
}

\twocolumn[{%
\renewcommand\twocolumn[1][]{#1}%
\maketitle

\begin{center}
    \centering
    \includegraphics[width=\linewidth]{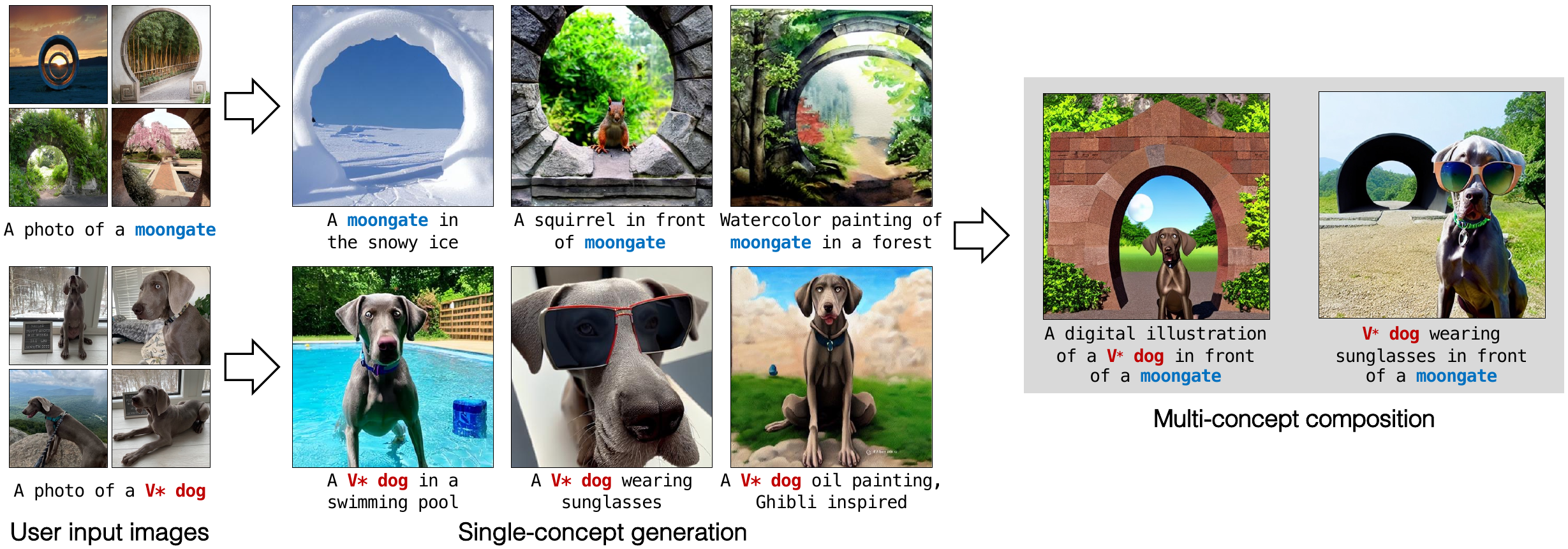}
    \vspace{-7pt}
\captionof{figure}{Given a few images of a new concept, our method augments a pre-trained text-to-image diffusion model, enabling new generations of the concept in unseen contexts. Example concepts include personal objects, animals, e.g., \textit{a pet dog}, and classes not well generated by the model, e.g., \textit{moongate} (a circular gate~\cite{moongatelink}). Furthermore, we propose a method for composing \textit{multiple} new concepts together, for example, {\menlo V$^*$ dog wearing sunglasses in front of a moongate}. We denote personal categories with a new modifier token {\menlo V$^*$}. }
    \label{fig:teaser}
\end{center}

}]
 \maketitle

\begin{abstract}
While generative models produce high-quality images of concepts learned from a large-scale database, a user often wishes to synthesize instantiations of their own concepts (for example, their family, pets, or items).
Can we teach a model to quickly acquire a new concept, given a few examples? Furthermore, can we compose multiple new concepts together? 
We propose Custom Diffusion, an efficient method for augmenting existing text-to-image models. We find that only optimizing a few parameters in the text-to-image conditioning mechanism is sufficiently powerful to represent new concepts while enabling fast tuning ($\smallsim6$ minutes). Additionally, we can jointly train for multiple concepts or combine multiple fine-tuned models into one via closed-form constrained optimization. Our fine-tuned model generates variations of multiple new concepts and seamlessly composes them with existing concepts in novel settings. Our method outperforms or performs on par with several baselines and concurrent works in both qualitative and quantitative evaluations while being memory and computationally efficient. 
\end{abstract}
\section{Introduction}
\lblsec{intro}

Recently released text-to-image models~\cite{ramesh2022hierarchical,saharia2022photorealistic,rombach2022high,yu2022scaling} have represented a watershed year in image generation. By simply querying a text prompt, users are able to generate images of unprecedented quality. Such systems can generate a wide variety of objects, styles, and scenes -- seemingly ``anything and everything''.

However, despite the diverse, \textit{general} capability of such models, users often wish to synthesize \textit{specific} concepts from their own personal lives. For example, loved ones such as family, friends, pets, or personal objects and places, such as a new sofa or a recently visited garden, make for intriguing concepts. As these concepts are by nature personal, they are unseen during large-scale model training. Describing these concepts after the fact, through text, is unwieldy and unable to produce the personal concept with sufficient fidelity. 

This motivates a need for model \textit{customization}. Given the few user-provided images, can we augment existing text-to-image diffusion models with the new concept (for example, their pet dog or a ``moongate'' as shown in \reffig{teaser})? The fine-tuned model should be able to generalize and compose them with existing concepts to generate new variations. This poses a few challenges -- first, the model tends to forget~\cite{ding2022don,ramasesh2021effect,li2022overcoming} or change~\cite{lee2019countering,lu2020countering} the meanings of existing concepts: e.g., the meaning of ``moon'' being lost when adding the ``moongate'' concept. Secondly, the model is prone to overfit the few training samples and reduce sampling variations.

Moreover, we study a more challenging problem, \emph{compositional fine-tuning} -- the ability to extend beyond tuning for a single, individual concept and compose multiple concepts together, e.g., {\menlo pet dog in front of moongate} (\reffig{teaser}). Improving compositional generation has been studied in recent works~\cite{liu2022compositional}. But composing multiple new concepts poses additional challenges, such as mixing unseen concepts. %

In this work, we propose a fine-tuning technique, \textit{Custom Diffusion} for text-to-image diffusion models. Our method is computationally and memory efficient. To overcome the above-mentioned challenges, we identify a small subset of model weights, namely the key and value mapping from text to latent features in the cross-attention layers~\cite{vaswani2017attention,bahdanau2014neural}. Fine-tuning these is sufficient to update the model with the new concept. To prevent model forgetting, we use a small set of real images with similar captions as the target images. We also introduce augmentation during fine-tuning, which leads to faster convergence and improved results. To inject multiple concepts, our method supports training on both simultaneously or training them separately and then merging.

We build our method on Stable Diffusion~\cite{stablediffusionlink} and experiment on various datasets with as few as four training images. For adding single concepts, our method shows better text alignment and visual similarity to the target images than concurrent works and baselines. More importantly, our method can compose multiple new concepts efficiently, whereas concurrent methods struggle and often omit one. Finally, our method only requires storing a small subset of parameters ($3\%$ of the model weights) and reduces the fine-tuning time (6 minutes on 2 A100 GPUs, $2-4 \times$ faster compared to concurrent works). Please find the code and data at our \href{https://www.cs.cmu.edu/~custom-diffusion/} {website}.
\section{Related Work}

\myparagraph{Deep generative models} Generative models aim to synthesize samples from a data distribution, given a set of training examples. These include GANs~\cite{goodfellow2020generative,stylegan3,biggan}, VAEs~\cite{kingma2013auto}, auto-regressive~\cite{esser2021taming,razavi2019generating,van2016pixel}, flow-based~\cite{dinh2014nice,dinh2016density}, and diffusion models~\cite{ho2020denoising,sohl2015deep,dhariwal2021diffusion}. To improve controllability, these models can be conditioned on a class~\cite{biggan,sauer2022stylegan}, image~\cite{isola2017image,wang2018pix2pixHD,zhu2017unpaired,meng2021sdedit}, or text prompt~\cite{zhu2019dm,tao2020df,nichol2021glide}. %
Our work mainly relates to text-conditioned synthesis~\cite{mansimov2015generating}.  While earlier works~\cite{reed2016generative, zhang2017stackgan,zhu2019dm,tao2020df,xu2018attngan,huang2022multimodal} were limited to a few classes~\cite{wah2011caltech,lin2014microsoft}, %
recent text-to-image models~\cite{nichol2021glide,saharia2022photorealistic,rombach2022high,ramesh2022hierarchical,dhariwal2021diffusion,ramesh2021zero,yu2022scaling}, trained on  extremely large-scale data, have demonstrated remarkable generalization ability. However, such models are by nature generalists and struggle to generate specific instances like personal toys or rare categories, e.g., ``moongate''. 
We aim to adapt these models to become specialists in new concepts.

\myparagraph{Image and model editing.} While generative models can sample random images, a user often wishes to edit a single, specific image. Several works aim at leveraging the capabilities of generative models, such as GANs~\cite{zhu2016generative,patashnik2021styleclip,abdal2019image2stylegan,abdal2020image2stylegan++,abdal2022clip2stylegan} or diffusion models~\cite{choi2021ilvr, nichol2021glide,kim2022diffusionclip} towards editing. A challenge is representing the specific image in the pre-trained model, and such methods employ per-image or per-edit optimization. A closely related line of work edits a generative model directly~\cite{bau2020rewriting,wang2022rewriting,gal2022stylegan}. Whereas these methods aim to customize GANs, our focus is on text-to-image models.

\myparagraph{Transfer learning.}
A method of efficiently producing a whole distribution of images is leveraging a pretrained model and then using transfer learning~\cite{wang2018TransferGAN,noguchi2019SB,wang2019MineGAN,mo2020FreezeD,zhao2020leveraging,li2020few,ojha2021few,liu2020towards,gu2021lofgan,nitzan2022mystyle,gal2022stylegan}. For example, one can adapt photorealistic faces into cartoons~\cite{noguchi2019SB,mo2020FreezeD,li2020few,ojha2021few,gal2022stylegan}. To adapt with just a few training images, efficient training techniques~\cite{stylegan2ada,zhao2020image,tran2020towards,diffaug,sauer2021projected,kumari2022ensembling} are often useful. Different from these works, which focus on tuning whole models to single domains, we wish to acquire multiple new concepts without catastrophic forgetting~\cite{french1999catastrophic,kirkpatrick2017overcoming,li2017learning,ramasesh2021effect,li2022overcoming}. By preserving the millions of concepts captured in large-scale pretraining, we can synthesize the new concepts in composition with these existing concepts. Relatedly, several methods~\cite{pfeiffer2020mad,skantze2022collie,gao2021clip,hu2021lora} propose to train adapter modules or low rank updates for large-scale models in the discriminative setting. In contrast, we adapt a small number of existing parameters and do not require additional parameters.

\myparagraph{Adapting text-to-image models.} Similar to our goals, two concurrent works, DreamBooth~\cite{ruiz2022dreambooth} and Textual Inversion~\cite{gal2022image}, adopt transfer learning to text-to-image diffusion models~\cite{saharia2022photorealistic,rombach2022high} via either fine-tuning all the parameters~\cite{ruiz2022dreambooth} or introducing and optimizing a word vector~\cite{gal2022image} for the new concept. Our work differs in several aspects. First, our work tackles a challenging setting: compositional fine-tuning of \textit{multiple} concepts, where concurrent works struggle. Second, we only fine-tune a subset of cross-attention layer parameters, which significantly reduces the fine-tuning time. We find these design choices lead to better results, validated by automatic metrics and human preference studies.

\begin{figure*}[!t]
    \centering
    \includegraphics[width=\linewidth]{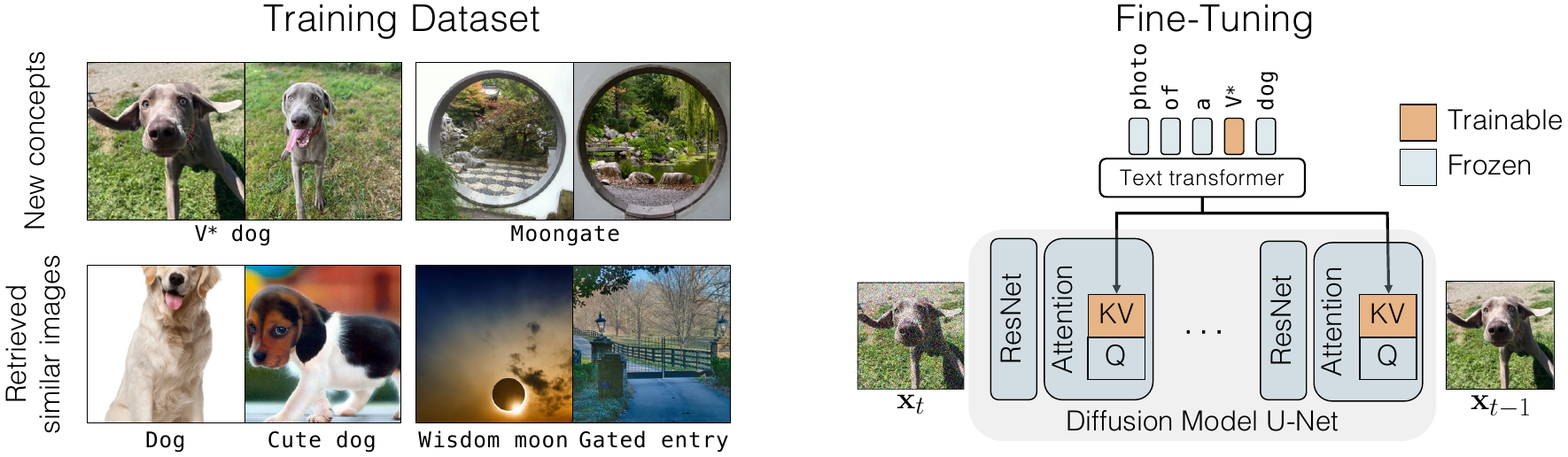}
    \caption{{\textbf{Custom Diffusion.} Given images of new concepts, we retrieve real images with similar captions as the given concepts and create the training dataset for fine-tuning, as shown on the left. To represent personal concepts of a general category, we introduce a new modifier token {\menlo V$^*$}, used in front of the category name. During training, we optimize key and value projection matrices in the diffusion model cross-attention layers along with the modifier token. The retrieved real images are used as a regularization dataset during fine-tuning.
    }}
    \lblfig{methoddiagram}
    \vspace{-15pt}
\end{figure*}

\section{Method}\label{sec:method}
Our proposed method for model fine-tuning, as shown in \reffig{methoddiagram}, only updates a small subset of weights in the cross-attention layers of the model. In addition, we use a regularization set of real images to prevent overfitting on the few training samples of the target concepts. In this section, we explain our design choices and final algorithm in detail.

\subsection{Single-Concept Fine-tuning}\label{sec:single_concept}

Given a pretrained text-to-image diffusion model, we aim to embed a new concept in the model given as few as four images and the corresponding text description. The fine-tuned model should retain its prior knowledge, allowing for novel generations with the new concept, based on the text prompt. This can be challenging as the updated text-to-image mapping might easily overfit the few available images. 

In our experiments, we use Stable Diffusion~\cite{stablediffusionlink} as our backbone model, which is built on the Latent Diffusion Model (LDM)~\cite{rombach2022high}. LDM first encodes images into a latent representation, using hybrid objectives of VAE~\cite{kingma2013auto}, PatchGAN~\cite{isola2017image}, and LPIPS~\cite{zhang2018unreasonable}, such that running an encoder-decoder can recover an input image. They then train a diffusion model~\cite{ho2020denoising} on the latent representation with text condition injected in the model using cross-attention.

\myparagraph{Learning objective of diffusion models.}
Diffusion models~\cite{sohl2015deep,ho2020denoising} are a class of generative models that aim to approximate the original data distribution $q(\x_0)$ with $p_\theta(\x_0)$:
\begin{equation}
    \begin{aligned}
    p_\theta(\x_0) %
    = \int \Bigr [ p_{\theta} (\x_{T}) \prod p_{\theta}^t(\x_{t-1} | \x_t) \Bigr ] d\x_{1:T}, 
    \end{aligned}
\end{equation}
where $\x_1$ to $\x_T$ are latent variables of a forward Markov chain s.t. $\x_{t} = \sqrt{\alpha_t}\x_{0} + \sqrt{1 - \alpha_t}\epsilon$. The model is trained to learn the reverse process of a fixed-length (usually $1000$) Markov chain. Given noisy image $\x_{t}$ at timestep $t$, the model learns to denoise the input image to obtain $\x_{t-1}$. The training objective of the diffusion model can be simplified to:   
\begin{equation}
    \begin{aligned}
     \mathbb{E}_{\epsilon,\x,\c, t } [w_t||\epsilon - \epsilon_{\theta} (\x_t, \c, t) ||], \\
    \end{aligned}\label{eq:loss}
\end{equation}

\noindent where $\epsilon_{\theta}$ is the model prediction and $w_t$ is a time-dependent weight on the loss. The model is conditioned on timestep $t$ and can be further conditioned on any other modality $\c$, e.g., text. During inference, a random Gaussian image (or latent) $\x_T$ is denoised for fixed timesteps using the model~\cite{song2020denoising}.

\begin{figure}[!t]
    \centering
    \includegraphics[width=0.9\linewidth]{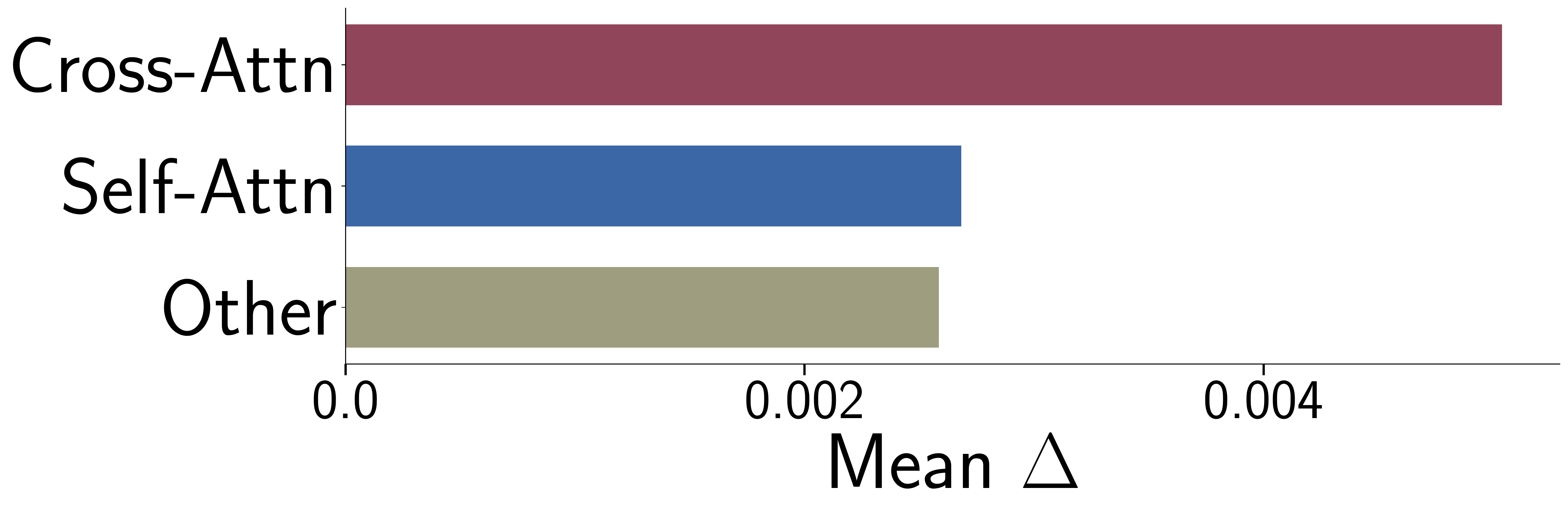}
    \vspace{-12pt}
    \caption{{\textbf{Analysis of change in weights} on updating all network weights during fine-tuning. The mean change in the cross-attention layers is significantly higher than other layers even though they only make up $5\%$ of the total parameter count.
    }}
    \lblfig{change}
    \vspace{-15pt}
\end{figure}

\noindent A na\"{i}ve baseline for the goal of fine-tuning is to update all layers to minimize the loss in Eqn.~\ref{eq:loss} for the given text-image pairs. This can be computationally inefficient for large-scale models and can easily lead to overfitting when training on a few images. Therefore, we aim to identify a minimal set of weights that is sufficient for the task of fine-tuning.

\myparagraph{Rate of change of weights.} Following Li~\etal~\cite{li2020few}, we analyze the change in parameters for each layer in the fine-tuned model on the target dataset with the loss in Eqn.~\ref{eq:loss}, i.e., $\Delta_l =  ||\theta_{l}' - \theta_{l}||/||\theta_{l}||,$
where $\theta_{l}'$ and $\theta_{l}$ are the updated and pretrained model parameters of layer $l$. These parameters come from three types of layers --  (1) cross-attention (between the text and image), (2) self-attention (within the image itself), and (3) the rest of the parameters, including convolutional blocks and normalization layers in the diffusion model U-Net. \reffig{change} shows the mean $\Delta_l$ for the three categories when the model is fine-tuned on ``moongate'' images. We observe similar plots for other datasets. As we see, the cross-attention layer parameters have relatively higher $\Delta$ compared to the rest of the parameters. Moreover, cross-attention layers are only $5\%$ of the total parameter count in the model. This suggests it plays a significant role during fine-tuning, and we leverage that in our method.

\myparagraph{Model fine-tuning.}
Cross-attention block modifies the latent features of the network according to the condition features, i.e., text features in the case of text-to-image diffusion models. Given text features $\c \in \mathbb{R}^{s \times d}$ and latent image features $\f \in \mathbb{R}^{ (h\times w) \times l}$, a single-head cross-attention~\cite{vaswani2017attention} operation consists of $Q = W^q \f, \;\; K = W^{k}\c, \;\; V = W^{v}\c$
, and a weighted sum over value features as:
\begin{equation}
    \begin{aligned}
    \text{Attention}(Q, K, V) = \text{Softmax}\Big(\frac{QK^T}{\sqrt{d'}} \Big)V, \\
    \end{aligned}\lbleq{objective}
\end{equation}

\noindent where $W^q$, $W^k$, and $W^v$ map the inputs to a query, key, and value feature, respectively, and $d'$ is the output dimension of key and query features. The latent feature is then updated with the attention block output. The task of fine-tuning aims at updating the mapping from given text to image distribution, and the text features are only input to $W^k$ and $W^v$ projection matrix in the cross-attention block. Therefore, we propose to only update $W^{k}$ and $W^{v}$ parameters of the diffusion model during the fine-tuning process. As shown in our experiments, this is sufficient to update the model with a new text-image paired concept. \reffig{caddiagram} shows an instance of the cross-attention layer and the trainable parameters.

\begin{figure}[!t]
    \centering
    \includegraphics[width=\linewidth]{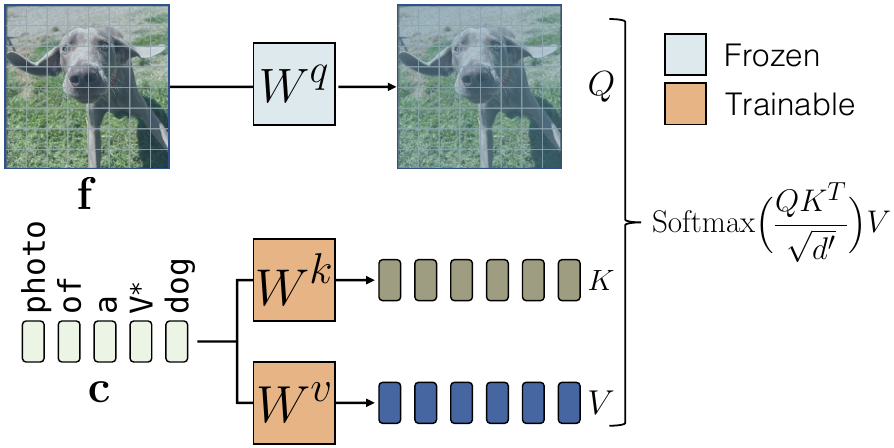}
    \vspace{-12pt}
    \caption{{\textbf{Single-head Cross-Attention.} Latent image feature $\f$ and text feature $\c$ are projected into query $Q$, key $K$, and value $V$. Output is a weighted sum of values, weighted by the similarity between the query and key features. We highlight the updated parameters $W^k$ and $W^v$ in our method.
    }}
    \lblfig{caddiagram}
    \vspace{-15pt}
\end{figure}

\myparagraph{Text encoding.}
Given target concept images, we require a text caption as well. If there exists a text description, e.g., {\menlo moongate}, we use that as a text caption. For personalization-related use-case where the target concept is a unique instance of a general category, e.g., pet dog, we introduce a new modifier token embedding, i.e., {\menlo V$^*$ dog}. During training, {\menlo V$^*$} is initialized with a rare occurring token embedding and optimized along with cross-attention parameters. An example text caption used during training is, {\menlo photo of a V$^{*}$ dog}. 

\myparagraph{Regularization dataset.}
Fine-tuning on the target concept and text caption pair can lead to the issue of language drift~\cite{lee2019countering,lu2020countering}. For example, training on  ``moongate'' will lead to the model forgetting the association of ``moon'' and ``gate'' with their previously trained visual concepts, as shown in \reffig{overfitting}. Similarly, training on a personalized concept of {\menlo V$^{*}$ tortoise plushy} can leak, causing all examples with {\menlo plushy} to produce the specific target images. To prevent this, we select a set of 200 regularization images from the LAION-400M~\cite{schuhmann2021laion} dataset with corresponding captions that have a high similarity with the target text prompt, above threshold $0.85$ in CLIP~\cite{radford2021learning} text encoder feature space. 

\begin{figure}[!t]
    \centering
    \includegraphics[width=\linewidth]{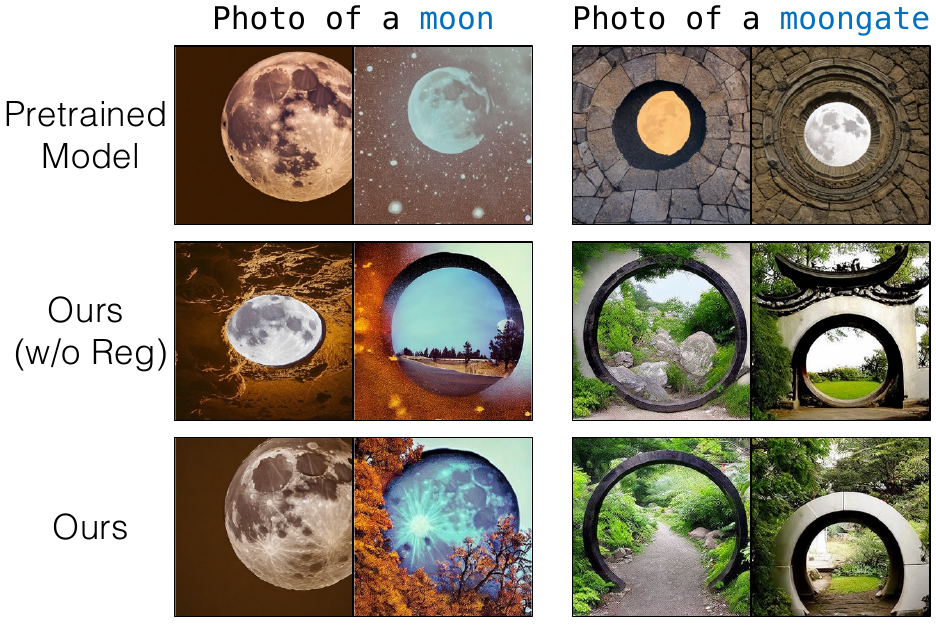}
    \vspace{-15pt}
    \caption{{\textbf{Role of regularization data in mitigating overfitting behavior during fine-tuning}. $1^{\text{st}}$ row: samples from pre-trained models. In $2^{\text{nd}}$ row, fine-tuning cross-attention key, value projection matrices without any regularization dataset leads to {\menlo moongate} like images on the text prompt {\menlo photo of a moon}. We largely mitigate this issue with the use of regularization datasets as shown in $3^{\text{rd}}$ row. More results can be found in \reffig{supp_overfit} (Appendix).
    }}
    \lblfig{overfitting}
    \vspace{-15pt}
\end{figure}
\subsection{Multiple-Concept Compositional Fine-tuning}\label{sec:multi-concept}

\myparagraph{Joint training on multiple concepts.}
For fine-tuning with multiple concepts, we combine the training datasets for each individual concept and train them jointly with our method. To denote the target concepts, we use different modifier tokens, {\menlo V$_i^*$}, initialized with different rarely-occurring tokens and optimize them along with cross-attention key and value matrices for each layer. As shown in \reffig{results_multi_concept}, restricting the weight update to cross-attention key and value parameters leads to significantly better results for composing two concepts compared to methods like DreamBooth, which fine-tune all the weights.

\myparagraph{Constrained optimization to merge concepts.}
As our method only updates the key and value projection matrices corresponding to the text features, we can subsequently merge them to allow generation with multiple fine-tuned concepts. Let set $\{ W^k_{0,l}, W^v_{0,l} \}_{l=1}^L$ represent the key and value matrices for all $L$ cross-attention layers in the pretrained model and $\{W^k_{n,l}, W^v_{n,l}\}_{l=1}^L$ represent the corresponding updated matrices for added concept $n \in \{1 \cdots N\}$. As our subsequent optimization applies to all layers and key-value matrices, we will omit superscripts $\{k, v\}$ and layer $l$ for notational clarity. We formulate the composition objective as the following constrained least squares problem: 
\begin{equation}
    \begin{split}
     \hat{W} = \argmin_{W} ||  W C_{\textnormal{reg}}^\top - & W_{0}C_{\textnormal{reg}}^\top ||_F \\
     \textnormal{s.t. } \hspace{1mm}  WC^\top = V, \textnormal{ whe}& \textnormal{re } C = [\c_1 \cdots \c_N]^\top \\
     \textnormal{and } & V = [W_{1}\c_1^\top \cdots W_{N}\c_N^\top]^\top.
    \end{split}\lbleq{formulation}
\end{equation}
\noindent Here, $C \in \mathbb{R}^{s \times d}$ is the text features of dimension $d$. These are compiled of $s$ target words across all $N$ concepts, with all captions for each concept flattened out and concatenated. Similarly, $C_{\textnormal{reg}} \in \mathbb{R}^{s_{\textnormal{reg}} \times d} $ consists of text features of $\smallsim1000$ randomly sampled captions for regularization. Intuitively, the above formulation aims to update the matrices in the original model, such that the words in target captions in $C$ are mapped consistently to the values obtained from fine-tuned concept matrices. The above objective can be solved in closed form, assuming $C_{\textnormal{reg}}$ is non-degenerate and the solution exists, by using the method of Lagrange multipliers~\cite{boyd2004convex}:
\begin{equation}
    \begin{split}
     \hat{W} =  W_{0} + \v^\top \d,
     \hspace{1mm} & \textnormal{where } \d = C(C_{\textnormal{reg}}^\top C_{\textnormal{reg}})^{-1} \\
     \textnormal{and } \v^\top & = ( V -  W_{0}C^\top )( \d C^\top )^{-1}.
    \end{split}\lbleq{solution}
\end{equation}
We show full derivation in the \refapp{derivation}. Compared to joint training, our optimization-based method is faster ($\smallsim2$ seconds) if each individual fine-tuned model exists. Our proposed methods lead to the coherent generation of two new concepts in a single scene, as shown in Section~\ref{sec:multiple_concept_expr}.

\myparagraph{Training details.}
We train with our method for $250$ steps in single-concept and $500$ steps in two-concept joint training, on a batch size of $8$ and learning rate $8 \times 10^{-5}$. During training, we also randomly resize the target images from $0.4 - 1.4 \times$ and append the prompt ``very small'', ``far away'' or ``zoomed in'', ``close up'' accordingly to the text prompt based on resize ratio. We only backpropagate the loss on valid regions. This leads to faster convergence and improved results. We provide more training details in the \refapp{details}.
\section{Experiments}\lblsec{expr}
In this section, we show the results of our method on multiple datasets in both single concept fine-tuning and composition of two concepts on the Stable Diffusion model~\cite{stablediffusionlink}. 

\myparagraph{Datasets.} We perform experiments on ten target datasets spanning a variety of categories and varying training samples. It consists of two scene categories, two pets, and six objects, as shown in \reffig{ind_values}. We also recently introduced a new dataset, CustomConcept101, consisting of 101 concepts, and show our results on it in \refapp{customconcept101} and the \href{https://www.cs.cmu.edu/~custom-diffusion/dataset.html} {website}.  

\myparagraph{Evaluation metrics.}
We evaluate our method on (1) \emph{Image-alignment}, i.e., the visual similarity of generated images with the target concept, using similarity in CLIP image feature space~\cite{gal2022image}, (2) \emph{Text-alignment} of the generated images with given prompts, using text-image similarity in CLIP feature space~\cite{hessel2021clipscore}, and (3) \emph{KID}~\cite{binkowski2018demystifying} on a validation set of $500$ real images of a similar concept retrieved from LAION-400M to measure overfitting on the target concept (e.g., V$^*$ dog)and forgetting of existing related concepts (e.g., dog). (4) We also perform a \emph{human preference} study of our method with baselines. Unless mentioned otherwise, we use $200$ steps of DDPM sampler with a scale $6$. The prompts used for quantitative and human evaluation are shown on our \href{https://www.cs.cmu.edu/~custom-diffusion/}{website}. 

\myparagraph{Baselines.}
We compare our method with the two concurrent works, \textit{DreamBooth}~\cite{ruiz2022dreambooth} (third-party implementation~\cite{dreamboothimpl}) and \textit{Textual Inversion}~\cite{gal2022image}. DreamBooth fine-tunes all the parameters in the diffusion model, keeping the text transformer frozen, and uses generated images as the regularization dataset. Each target concept is represented by a unique identifier, followed by its category name, i.e., ``[V] \textit{category}'', where [V] is a rarely occurring token in the text token space and not optimized during fine-tuning. Textual Inversion optimizes a new {\menlo V$^*$} token for each new concept. We also compare with the competitive baseline of \textit{Custom Diffusion (w/ fine-tune all)}, where we fine-tune all the parameters in the U-Net~\cite{ronneberger2015u} diffusion model, along with the {\menlo V$^*$} token embedding in our method. We provide implementation details for all baselines in the supplement.
\vspace{-5pt}

\subsection{Single-Concept Fine-tuning Results}\label{sec:single_concept_results}

\begin{figure*}[!t]
    \centering
    \includegraphics[width=\linewidth]{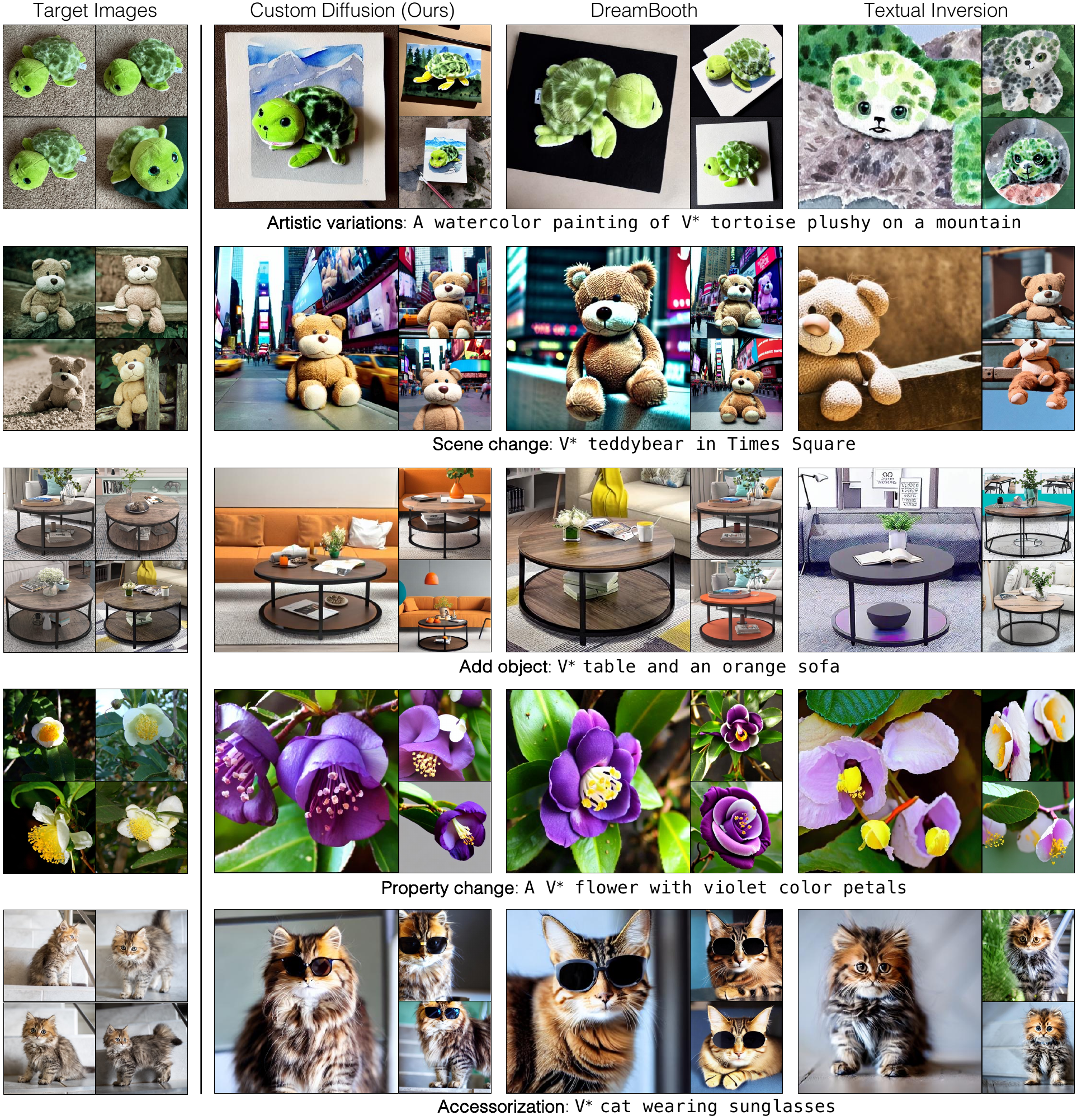}
    \vspace{-10pt}
    \caption{{\textbf{Single-concept fine-tuning results.} Given images of a new concept (target images shown on the left), our method can generate images with the concept in unseen contexts and art styles. \textit{\underline{First row}}: representing the concept in an artistic style of watercolor paintings. Our method can also generate the mountains in the background, which DreamBooth and Textual Inversion omit. \textit{\underline{Second row}}: changing the background scene. Our method and DreamBooth perform similarly and better than Textual Inversion. \textit{\underline{Third row}}: adding another object, e.g., an orange sofa with the target table. Our method successfully adds the other object. \textit{\underline{Fourth row}}: changing object property, like color of petals. \textit{\underline{Fifth row}}: accessorizing personal pet cat with sunglasses. Our method preserves the visual similarity better than baselines while changing only the petal colors or adding sunglasses to the cat. For Textual Inversion, the input text prompt consists of only optimized {\menlo V$^*$} instead of {\menlo V$^*$ category}. We show more samples on our \href{https://www.cs.cmu.edu/~custom-diffusion/}{website}.
    } }
    \label{fig:results_single_cat}
    \vspace{-5pt}
\end{figure*}

\myparagraph{Qualitative evaluation.}
We test each fine-tuned model on a set of challenging prompts. This includes generating the target concept in a new scene, in a known art style, composing it with another known object, and changing certain properties of the target concept: e.g., color, shape, or expression. Figure~\ref{fig:results_single_cat} shows the sample generations with our method, DreamBooth, and Textual Inversion. Our method, Custom Diffusion, has higher text-image alignment while capturing the visual details of the target object. It performs better than Textual Inversion and is on par with DreamBooth while having a lower training time and model storage ($\smallsim5\times$ faster and $75$MB vs $3$GB storage).

\myparagraph{Quantitative evaluation.}
We evaluate on $20$ text prompts and $50$ samples per prompt for each dataset, resulting in a total of $1000$ generated images. We use DDPM sampling with $50$ steps and a classifier-free guidance scale of $6$ across all methods. As shown in \reffig{ind_values}, our method outperforms the concurrent methods~\cite{ruiz2022dreambooth,gal2022image}. Also, Custom Diffusion works on par with our proposed baseline of fine-tuning all the weights in the diffusion model, while being more computationally and time efficient. \reftbl{results} also shows the KID of generated images by each fine-tuned model on a reference dataset, with captions similar to the fine-tuned concept. As we observe, our method has lower KID than most baselines, which suggests less overfitting to the target concept. In \refapp{exprmore}, we show that the updated matrices can be compressed to further reduce model storage.

\myparagraph{Computational requirements}Training time of our method is $\smallsim6$ minutes (2 A100 GPUs), compared to $20$ minutes for Ours (w/ fine-tune all) (4 A100s), $20$ minutes for Textual Inversion (2 A100s), and $\smallsim1$ hour for DreamBooth (4 A100s). Also, since we update only $75$MB of weights, our method has low memory requirements for storing each concept model. We keep the batch size fixed at $8$ across all.

\begin{figure*}[!t]
    \centering
    \includegraphics[width=\linewidth]{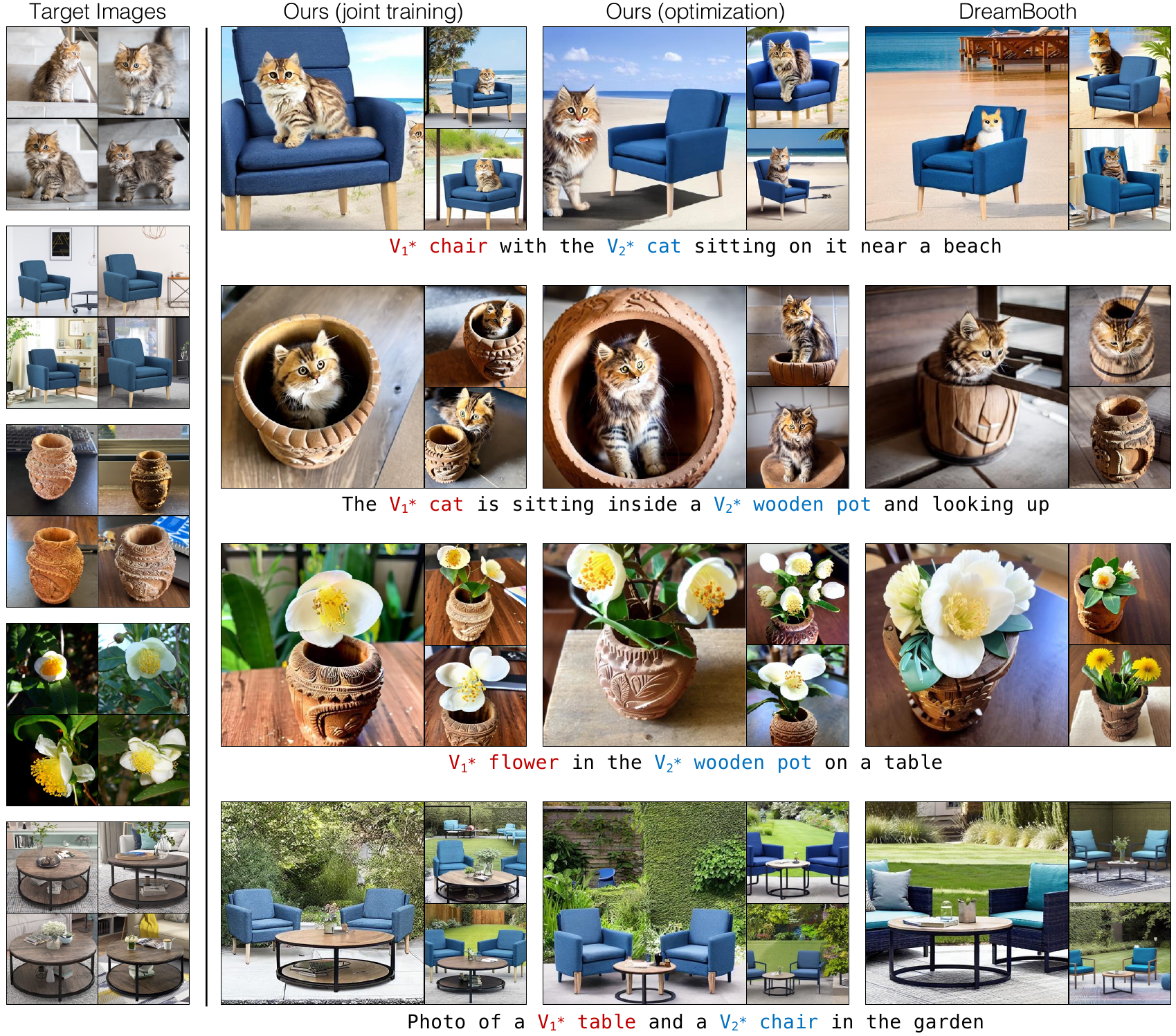}
    \vspace{-20pt}
    \caption{{\textbf{Multi-concept fine-tuning results.} \textit{\underline{First row}}: our method has higher visual similarity with the personal cat and chair images shown in the first column while following the text condition. \textit{\underline{Second row}}: DreamBooth omits the cat in 3 out of 4 images, whereas our method generates both cats and wooden pots. \textit{\underline{Third row}}: our method generates the target flower in the wooden pot while maintaining the visual similarity to the target images. \textit{\underline{Fourth row}}: generating the target table and chair together in a garden. For all settings, our optimization-based approach and joint training perform better than DreamBooth, and joint training performs better than the optimization-based method. 
    }}
    \label{fig:results_multi_concept}
    \vspace{-15pt}
\end{figure*}

\subsection{Multiple-Concept Fine-tuning Results}\label{sec:multiple_concept_expr}
We show the results of generating two new concepts in the same scene on the following five pairs: (1) Moongate + Dog, (2) Cat + Chair, (3) Wooden Pot + Cat, (4) Wooden Pot + flower, and (5) Table + Chair. We compare our method with DreamBooth training on the two datasets together, using two different $[V_1]$ and $[V_2]$ tokens for each concept. For Textual Inversion, we perform inference using the individually fine-tuned tokens for each concept in the same sentence. We compare our method with the baselines on a set of $400$ images generated with $8$ prompts for each composition setting in \reffig{ind_values} and \reftbl{results}. We also compare with our baseline of sequentially training on the two concepts or fine-tuning all weights in our method. Our method performs better on all except the ``Table+Chair'' composition, where all methods perform comparably except Textual Inversion, which doesn't perform well at composition as also shown in \reffig{supp_ti_compose} in the Appendix. This shows the importance of fine-tuning only the cross-attention parameters for composition. In the case of sequential training, we observe forgetting of the first concept. Figure~\ref{fig:results_multi_concept} shows sample images of our proposed two methods and DreamBooth. As we can see, our method is able to generate the two objects in the same scene in a coherent manner while having high alignment with the input text prompt. We show more samples on our \href{https://www.cs.cmu.edu/~custom-diffusion/}{website}.
\vspace{-5pt}

\begin{figure*}[!t]
    \centering
    \includegraphics[width=\linewidth]{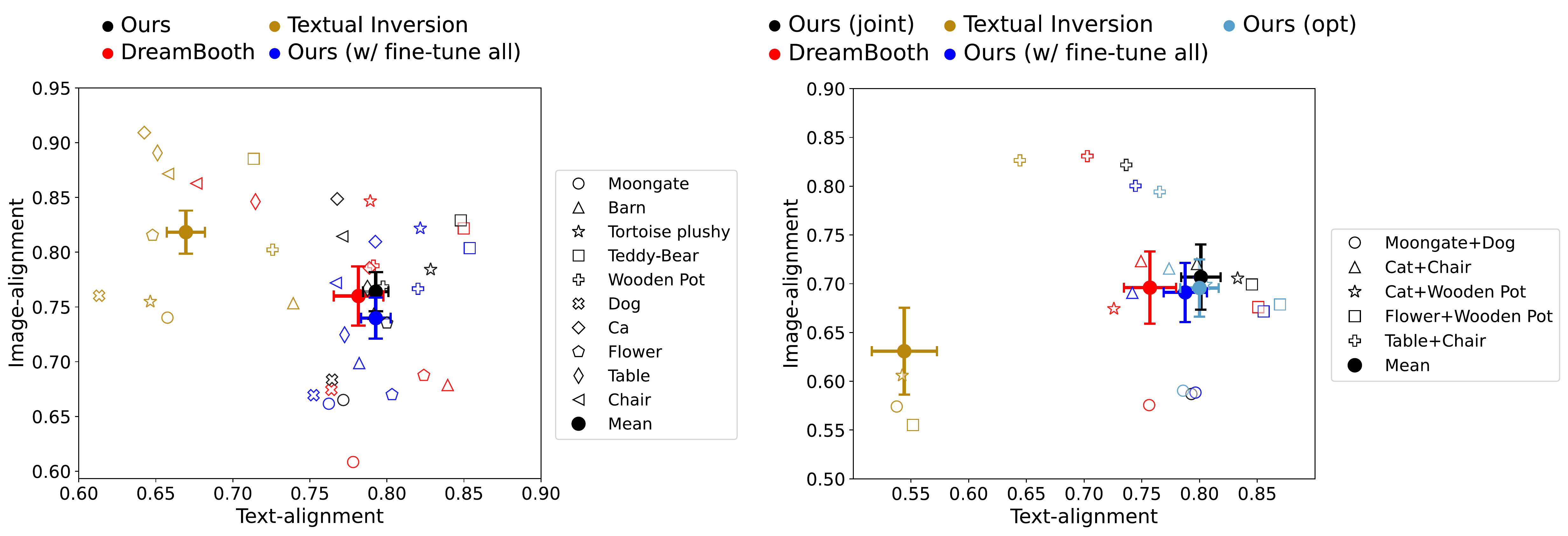}
    \vspace{-20pt}
    \caption{{\textbf{Text- and image-alignment for single-concept (left) and multi-concept (right) fine-tuning}. Compared to DreamBooth, Textual Inversion, and our w/ fine-tune all baseline, our method lies further along the upper right corner (with less variance). There exists a trade-off between high similarity to target images vs. text-alignment on new prompts. Keeping in consideration this trade-off, our method is on-par or better than the baselines. For multi-concept both joint training and optimization based method are better than other baselines.
    }}
    \lblfig{ind_values}
    \vspace{-15pt}
\end{figure*}

\begin{table}[!t]
\centering
\setlength{\tabcolsep}{5pt}
\resizebox{\linewidth}{!}{
\begin{tabular}{ll c c  c }
\toprule
& \textbf{Method}
& \shortstack[c]{\textbf{Text-alignment} } 
& \shortstack[c]{\textbf{Image-alignment} }
& \shortstack[c]{\textbf{KID (validation)}} \\
\midrule
\multirow{4}{*}{\shortstack[c]{\textbf{Single-}\\ \textbf{Concept} } }  & Textual Inversion    & 0.670 & \textbf{0.827} &  22.27  \\
& DreamBooth & 0.781   & 0.776 & 32.53 \\
\cdashline{2-5}
& Ours (w/ fine-tune all)   & \textbf{0.795} & 0.748 & \textbf{19.27}  \\
& Ours    & \textbf{0.795} & 0.775 & 20.96 \\
\midrule

\multirow{6}{*}{\shortstack[c]{\textbf{Multi-}\\ \textbf{Concept} } } 
&   Textual Inversion   & 0.544 & 0.630 & \multirow{6}{*}{---} \\
& DreamBooth & 0.783 & 0.695 &  \\
\cdashline{2-5}
& Ours (w/ fine-tune all) & 0.787 & 0.691 &  \\
& Ours (Sequential) & 0.797 & 0.700 &  \\
& Ours (Optimization)& 0.800 & 0.695 &  \\
& Ours (Joint)    & \textbf{0.801} & \textbf{0.706} &  \\ 
\bottomrule
\vspace{-10pt}
\end{tabular}
}
\vspace{-8pt}
\caption{\textbf{Quantitative comparisons.}  \textit{\underline{Top row}}: single-concept fine-tuning evaluation averaged across datasets. The last column shows the KID ($\times 10^3$) between real validation set images and generated images with the same caption. Since our method uses a regularization set of real images, it achieves lower KID and even improves slightly over the pretrained model. Textual Inversion has the same KID as the pretrained model, as it does not update the model.  \textit{\underline{Bottom row}}: evaluation on multi-concept averaged across the five composition pairs. We show individual scores for all in \reftbl{ind_values_single_concept} and \ref{tbl:supp_multi-concept-training_eval} in the Appendix. We also evaluate single-concept fine-tuned models on FID~\cite{fid} (MS-COCO~\cite{lin2014microsoft}) in \reftbl{mscoco} and show the trend of image-, text-alignment with training steps in \reffig{training_iter}.}

\label{tbl:results}
\vspace{-6pt}
\end{table}

\begin{table}[!t]
\centering
\setlength{\tabcolsep}{5pt}
\resizebox{\linewidth}{!}{
\begin{tabular}{@{\extracolsep{4pt}}ll c  cc cc@{} }
\toprule

\textbf{Ours}   
&  \multicolumn{2}{@{} c}{\textbf{Textual Inversion}} 
&  \multicolumn{2}{@{} c}{\textbf{DreamBooth}} 
&  \multicolumn{2}{@{} c}{\textbf{Ours (w/ fine-tune all)}}  \\

\cmidrule{2-3} \cmidrule{4-5} \cmidrule{6-7}  
 &  \multirow{2}{*}{\shortstack[c]{Text\\ Alignment }}
&  \multirow{2}{*}{\shortstack[c]{Image\\ Alignment} }
& \multirow{2}{*}{\shortstack[c]{Text\\ Alignment }  }
& \multirow{2}{*}{\shortstack[c]{Image\\ Alignment} }
& \multirow{2}{*}{\shortstack[c]{Text\\ Alignment}  }
& \multirow{2}{*}{\shortstack[c]{Image\\ Alignment} }\\ \\ 
\midrule

\multicolumn{1}{@{} l}{ \multirow{2}{*}{\textbf{\shortstack[c]{Single-concept}}} } &  \multirow{2}{*}{\shortstack[c]{ \textbf{72.62} \\ $\pm$ 2.38$\%$ }}  & \multirow{2}{*}{\shortstack[c]{ \textbf{51.62} \\ $\pm$ 2.62$\%$ }}  &
\multirow{2}{*}{\shortstack[c]{ \textbf{53.50} \\ $\pm$ 2.64$\%$ }} &
\multirow{2}{*}{\shortstack[c]{ \textbf{56.62} \\ $\pm$ 2.44$\%$ }} & 
\multirow{2}{*}{\shortstack[c]{ \textbf{55.17} \\ $\pm$ 2.55$\%$ }} &
\multirow{2}{*}{\shortstack[c]{ \textbf{53.99} \\ $\pm$ 2.44$\%$ }} \\ \\ 

\midrule

\multirow{2}{*}{\shortstack[c]{\textbf{Multi-concept}\\ (Joint)} } &

\multirow{2}{*}{\shortstack[c]{ \textbf{86.65} \\ $\pm$ 2.25 $\%$ }} &
\multirow{2}{*}{\shortstack[c]{ \textbf{81.89 } \\ $\pm$ 2.09 $\%$ }} &
\multirow{2}{*}{\shortstack[c]{ \textbf{56.39} \\ $\pm$ 2.46 $\%$ }} &  
\multirow{2}{*}{\shortstack[c]{ \textbf{61.80} \\ $\pm$ 2.59$\%$ }} & 
\multirow{2}{*}{\shortstack[c]{ \textbf{59.00} \\ $\pm$ 2.61 $\%$  }} & 
\multirow{2}{*}{\shortstack[c]{ \textbf{59.12}  \\ $\pm$ 2.72$\%$  }}  \\ \\

\cdashline{2-7}
\multirow{2}{*}{\shortstack[c]{\textbf{Multi-concept}\\ (Optimization)} } &

\multirow{2}{*}{\shortstack[c]{ \textbf{81.22} \\ $\pm$ 2.72$\%$ }} &
\multirow{2}{*}{\shortstack[c]{ \textbf{83.11} \\ $\pm$ 2.18$\%$ }} &
\multirow{2}{*}{\shortstack[c]{ \textbf{57.00} \\ $\pm$ 2.62 $\%$ }} &  
\multirow{2}{*}{\shortstack[c]{ \textbf{61.75} \\ $\pm$ 2.68$\%$ }} &
\multirow{2}{*}{\shortstack[c]{ \textbf{57.60} \\ $\pm$ 2.43 $\%$  }} & 
\multirow{2}{*}{\shortstack[c]{ \textbf{53.49}  \\ $\pm$ 2.71$\%$  }}  \\ \\

\bottomrule
\vspace{-10pt}
\end{tabular}
}
\vspace{-8pt}
\caption{\textbf{Human preference study.} For each paired comparison, our method is preferred (over $\geq 50\%$) over the baseline in both image- and text-alignment. Textual Inversion seems to overfit to target images and thus has a similar image-alignment as ours but performs worse on text-alignment in the single-concept setting.
}

\label{tbl:human_eval}
\vspace{-18pt}
\end{table}

\subsection{Human Preference Study}
We perform the human preference study using Amazon Mechanical Turk. We do a paired test of our method with DreamBooth, Textual Inversion, and Ours (w/ fine-tune all). For text-alignment, we show the two generations from each method (ours vs. baseline) on the same prompt with the question -- ``Which image is more consistent with the text?''. In image-alignment, we show 2-4 training samples of the relevant concept along with the generated images (same as in text-alignment study) with the question -- ``Which image better represents the objects as shown in target images?''. We collect $800$ responses for each questionnaire. As shown in \reftbl{human_eval}, our method is preferred over baselines in both single-concept and multi-concept, even compared to Ours (w/ fine-tune all) method, which shows the importance of only updating cross-attention parameters.

\begin{table}[!t]
\centering
\setlength{\tabcolsep}{5pt}
\resizebox{\linewidth}{!}{
\begin{tabular}{l c c  c }
\toprule
\textbf{Method}
& \textbf{Text-alignment}
& \textbf{Image-alignment}
& \textbf{KID (validation)} \\
\midrule
\textbf{Ours}    & 0.795 & \textbf{0.775} &  \textbf{20.96}  \\
\textbf{Ours w/o Aug} & \textbf{0.800}   & 0.736 & 20.67 \\
\textbf{Ours w/o Reg}   & 0.799 & 0.756 & 32.64  \\
\textbf{Ours w/ Gen}    & 0.791 & 0.768 & 34.70 \\
\bottomrule
\vspace{-10pt}
\end{tabular}
}
\vspace{-8pt}
\caption{\textbf{Ablation Study.} No augmentation during training leads to lower image-alignment. No regularization dataset or using generated images as regularization produces much worse KID. 
}

\label{tbl:ablation}
\vspace{-18pt}
\end{table}

\begin{figure*}[!t]
    \centering
    \includegraphics[width=\linewidth]{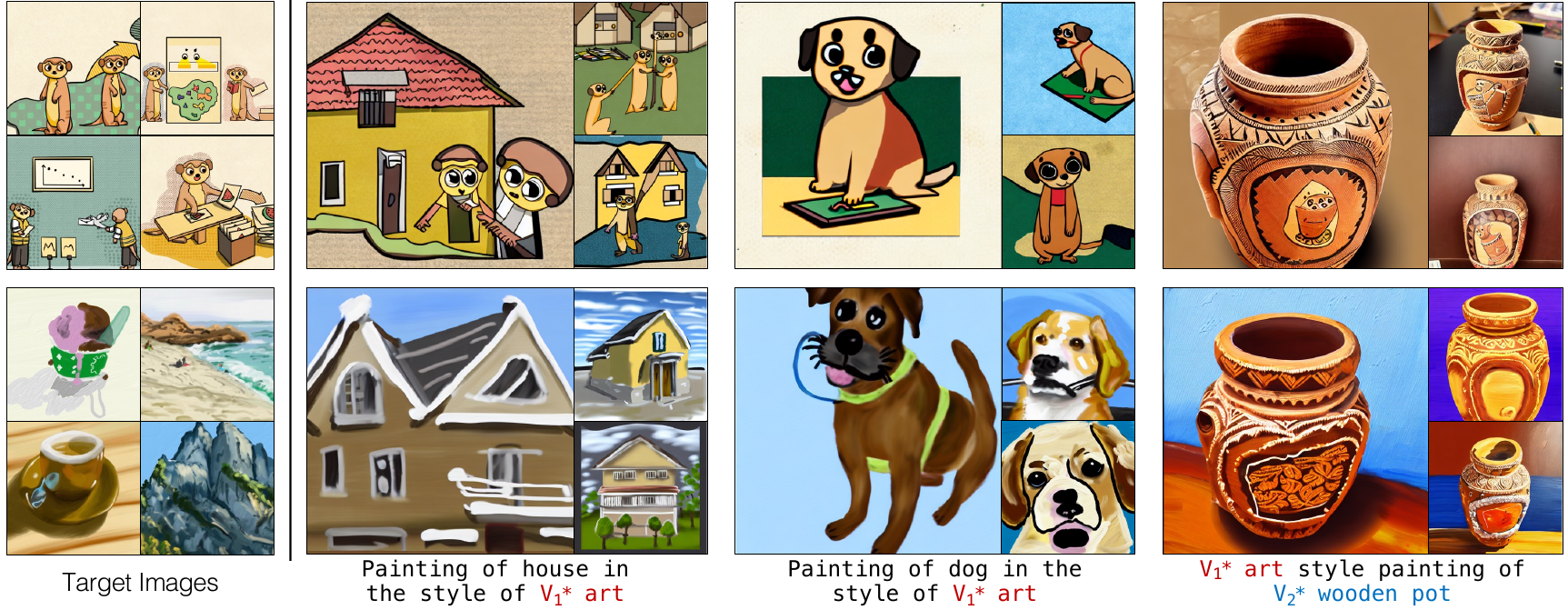}
    \vspace{-20pt}
    \caption{{\textbf{Custom Diffusion with artistic styles.} Our method can also be used to learn new styles. We finetune our model on 13 and 15 images for the top and bottom rows, respectively. The last column shows the composition of style with the new ``wooden pot'' concept (target images shown in \reffig{results_multi_concept}) using the optimization-based method. Style image credits: \href{https://www.mia-tang.com/blog}{Mia Tang} (top row), \href{https://www.instagram.com/aaronhertzmann/?hl=en}{Aaron Hertzmann} (bottom row). 
    }}
    \label{fig:supp_style}
    \vspace{-15pt}
\end{figure*}

\subsection{Ablation and Applications}\lblsec{ablation}
In this section, we ablate various components of our method to show its contribution. We evaluate each experiment on the same setup as in Section \ref{sec:single_concept_results}. We show sample generations for ablation experiments on our \href{https://www.cs.cmu.edu/~custom-diffusion/}{website}.

\myparagraph{Generated images as regularization (Ours w/ Gen).}
As detailed in Section~\ref{sec:single_concept}, we retrieve similar category real images and captions to use as regularization during fine-tuning. Another way of creating the regularization dataset is to generate images from the pretrained model~\cite{ruiz2022dreambooth}. We compare our method with this setup, i.e., for the target concept of a ``\textit{category}'' generate images using the prompt, {\menlo photo of a \{category\}}, and show results in Table~\ref{tbl:ablation}. Using generated images results in a similar performance on the target concept. However, this shows signs of overfitting, as measured by KID on a validation set of similar category real images.
\begin{figure}[!t]
    \centering
    \includegraphics[width=\linewidth]{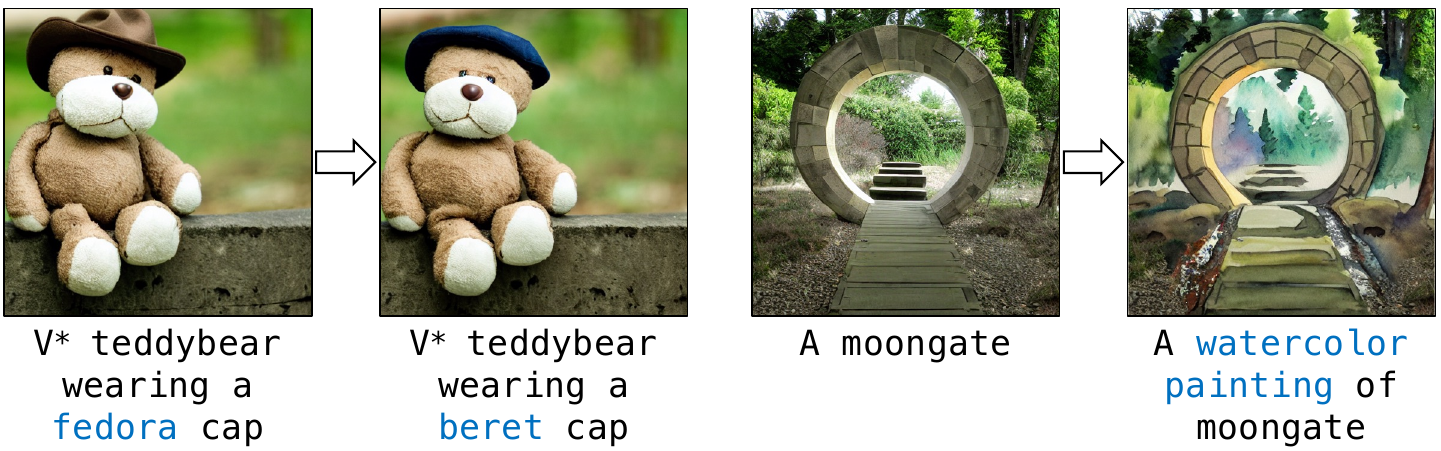}
    \vspace{-20pt}
    \caption{{\textbf{Image editing with our models using Prompt-to-Prompt~\cite{hertz2022prompt}.} \textit{Left:} replacement edit to only change the cap. \textit{Right:} image stylization while preserving the image content. 
    }}
    \label{fig:application}
    \vspace{-8pt}
\end{figure}

\begin{figure}[!t]
    \centering
    \includegraphics[width=\linewidth]{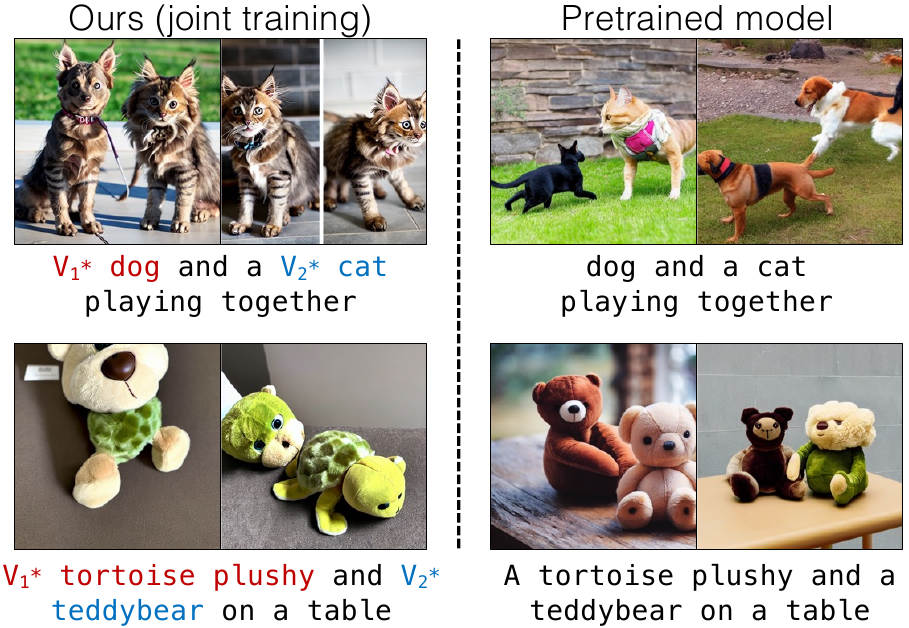}
    \vspace{-18pt}
    \caption{{\textbf{Failure cases on multi-concept fine-tuning.} Our method fails at difficult compositions like a cat and dog together in a scene or similar category objects like teddybear and tortoise plushy as shown on the right. Though as shown on the left, the pretrained model also struggles with similar compositions.
    }}
    \label{fig:results_multi_concept_limitation}
    \vspace{-18pt}
\end{figure}

\myparagraph{Without regularization dataset (Ours w/o Reg).}
We show results when no regularization dataset is used. We train the model for half the number of iterations (the same number of target images seen during training). Table~\ref{tbl:ablation} shows that the model has slightly lower image-alignment and tends to forget existing concepts, as evident from high KID on the validation set.

\myparagraph{Without data augmentation (Ours w/o Aug).}
As mentioned in Section~\ref{sec:multi-concept}, we augment by randomly resizing the target images during training and append size-related prompts (e.g., ``very small'') to the text. Here, we show the effect of not using these augmentations. The model is trained for the same number of steps. Table~\ref{tbl:ablation} shows that no augmentation leads to lower visual similarity with target images.

\myparagraph{Fine-tuning on a style.}
We show in \reffig{supp_style} that our method can also be used for fine-tuning with specific styles. We change the text prompt for target images to {\menlo A painting in the style of V$^*$ art}~\cite{gal2022image}. The regularization images are retrieved with captions having high similarity with ``art''.

\myparagraph{Image editing with fine-tuned models}
Similar to the pretrained model, our fine-tuned models can be used by existing image-editing methods. We show an example of using the Prompt-to-prompt~\cite{hertz2022prompt} method in \reffig{application}. 

\vspace{-8pt}
\vspace{5pt}
\section{Discussion and Limitations}
In conclusion, we have proposed a method for fine-tuning large-scale text-to-image diffusion models on new concepts, categories, personal objects, or artistic styles, using just a few image examples. Our computationally efficient method can generate novel variations of the fine-tuned concept in new contexts while preserving the visual similarity with the target images. Moreover, we only need to save a small subset of model weights. Furthermore, our method can coherently compose multiple new concepts in the same scene.

As shown in Figure \ref{fig:results_multi_concept_limitation}, difficult compositions, e.g., a pet dog and a pet cat, remain challenging. In this case, the pre-trained model also faces a similar difficulty, and our model inherits these limitations. Additionally, composing increasing three or more concepts together is also challenging. We show more analysis and visualization in the \refapp{exprmore}. 
\clearpage 

\myparagraph{Acknowledgment.}
We are grateful to Nick Kolkin, David Bau, Sheng-Yu Wang, Gaurav Parmar, John Nack, and Sylvain Paris for their helpful comments and discussion and to Allie Chang, Chen Wu, Sumith Kulal, Minguk Kang, Yotam Nitzan, and Taesung Park for proofreading the draft. We also thank Mia Tang and Aaron Hertzmann for sharing their artwork. We also appreciate the help of Sheng-Yu Wang, Songwei Ge, Daohan Lu, Ruihan Gao, Roni Shechtman, Avani Sethi, Yijia Wang, Shagun Uppal, and Zhizhuo Zhou in collecting the \href{https://www.cs.cmu.edu/~custom-diffusion/dataset.html}{dataset}, and Nick Kolkin for the feedback regarding it. This work was partly done by Nupur Kumari during the Adobe internship. The work is partly supported by Adobe Inc.

{\small
\bibliographystyle{ieee_fullname}
\bibliography{main}
}

\appendix
\renewcommand{\thefootnote}{\arabic{footnote}}

\clearpage
\noindent{\Large\bf Appendix}
\vspace{5pt}

In \refsec{customconcept101}, we show the results of our method on the CustomConcept101 dataset, which consists of a significantly larger number of custom concepts from a wide variety of categories. In \refsec{derivation}, we show the derivation of our optimization-based method for merging multiple concepts into a single model. In \refsec{exprmore}, we show additional experiment results of our method. We then provide more evaluation results, for example, the trend with training iterations, in \refsec{evalmore} and implementations details in \refsec{details}. Finally, we discuss the societal implications of our method in \refsec{society}.

\section{CustomConcept101}
\begin{figure*}[!t]
    \centering
    \includegraphics[width=\linewidth]{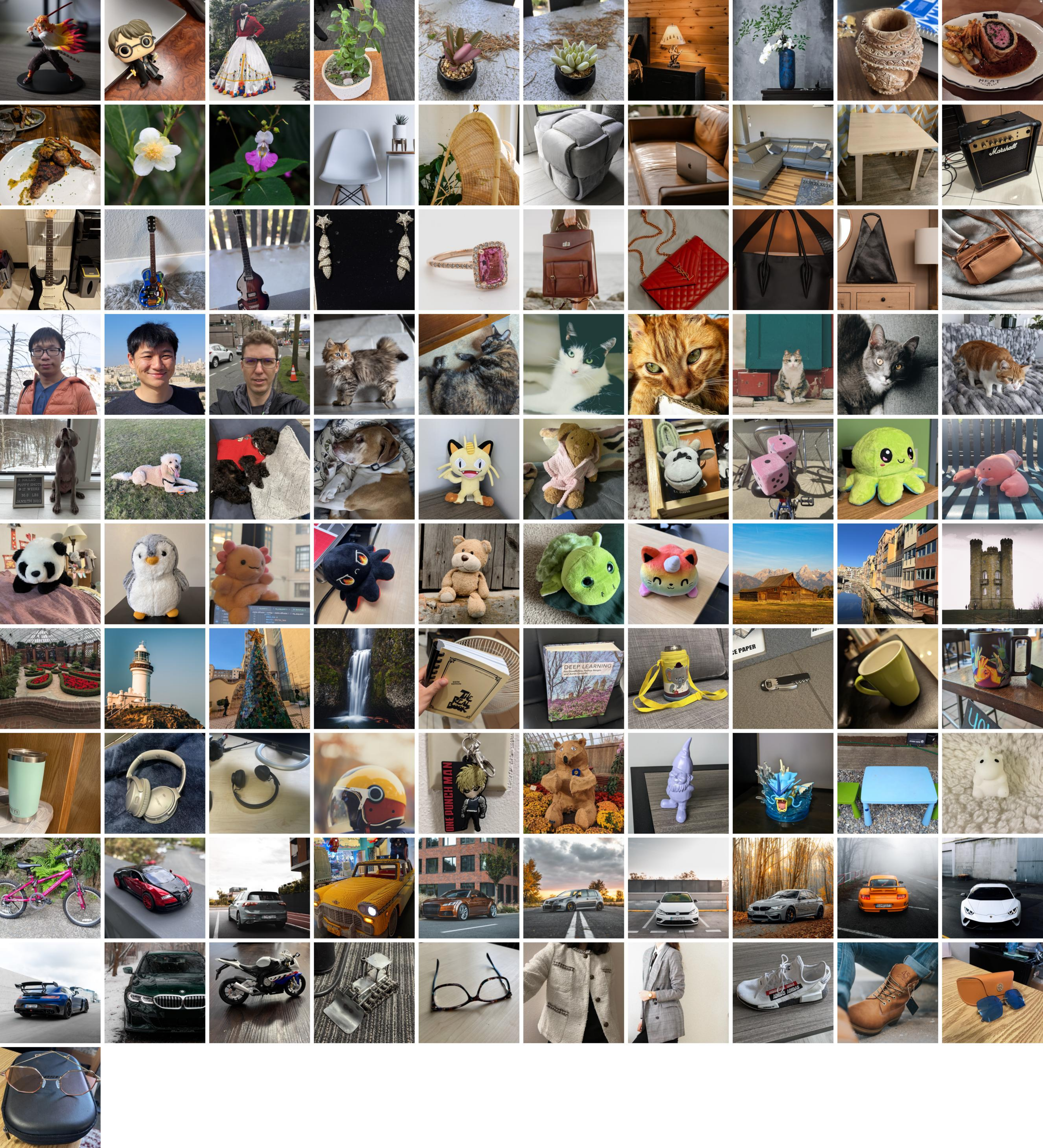}
    \vspace{-20pt}
    \caption{{\textbf{CustomConcept101 dataset.} A example image of each concept in the dataset. Out of $101$ categories, $31$ are collected from Unsplash, $2$ concepts belonging to the flower category are collected from other websites which allow redistribution, and the rest are captured by ourselves. For more details regarding the dataset and license, please refer to our \href{https://www.cs.cmu.edu/~custom-diffusion/dataset.html} {webpage}. 
    }}
    \label{fig:customconcept101}
    \vspace{-7pt}
\end{figure*}

\lblsec{customconcept101}
We show the results on our newly released dataset of 101 concepts consisting of a wide variety of categories, including toys, plushies, wearables, pets, scenes, and human faces. \reffig{customconcept101} shows one image of each concept from the dataset.  The images are collected from websites that allow redistribution, e.g., Unsplash, or captured by ourselves. For multi-concept, we propose $101$ unique compositions among the concepts, e.g., {\menlo V$_1^*$ dog with V$_2^*$ sunglasses}. We also create text prompts for evaluating the customization, $20$ prompts each in the case of single-concepts, and $12$ prompts for multi-concept composition. We first used ChatGPT~\cite{chatgpt} to propose $40$ prompts consisting of a particular concept or pair of concepts. We instruct it to change the background scene, insert another object in the scene, or create stylistic variations of the main subjects. We then manually filtered and modified the proposals to create a final set of prompts. 

\begin{figure*}[!t]
    \centering
    \includegraphics[width=\linewidth]{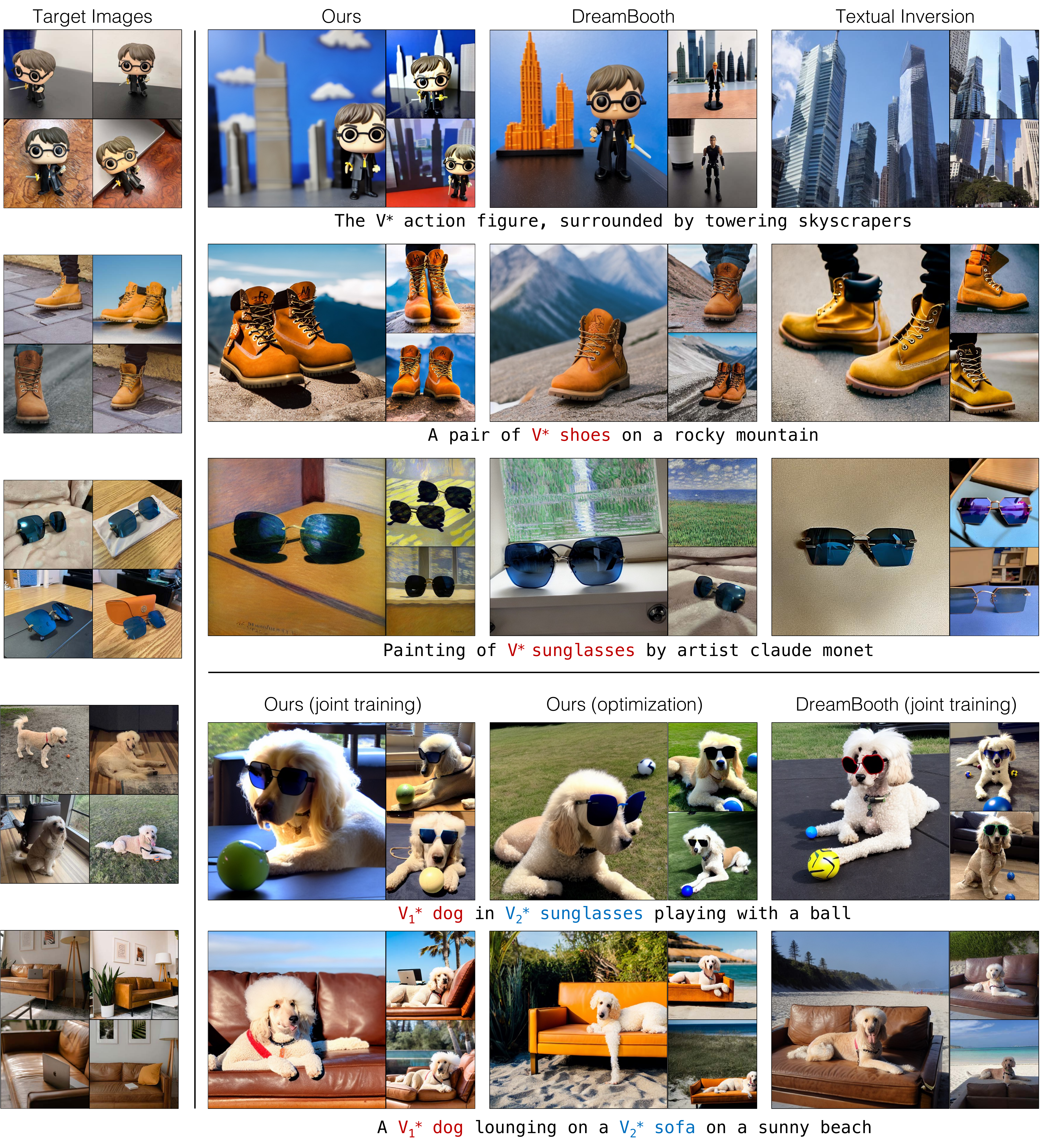}
    \vspace{-20pt}
    \caption{{\textbf{Qualitative comparison on CustomConcept101.} \textit{\underline{Single-concept}} top block (first three rows) shows sample comparisons of our method with DreamBooth and Textual Inversion on three concepts from the dataset. Our method shows higher text alignment, whereas DreamBooth retains more image alignment with the target images. \textit{\underline{Multi-concept}}: bottom block (last two rows) shows multi-concept composition using our method and DreamBooth. Our joint training method outperforms it on image-alignment with both target images while maintaining text-alignment with the given caption. The optimization method fails to preserve the image similarity in some cases, e.g., the sofa in the last row. We select the best three out of five images for single-concept and the best three out of 10 images for multi-concept. 
    }}
    \label{fig:results_customconcept101}
\end{figure*}

\reftbl{results_customconcept101} shows the comparison of our method with DreamBooth and Textual Inversion. Both ours and DreamBooth models are trained with generated images as regularization on this dataset. In the single-concept case, our method performs on par with DreamBooth with marginally lower image-alignment and higher text-alignment. Textual Inversion overfits the target images and has low text-alignment. For multi-concept customization, both our optimization and joint training method are better than DreamBooth on average. Sample images for each method are shown in \reffig{results_customconcept101}. For more results and dataset details, please refer to our \href{https://github.com/adobe-research/custom-diffusion/customconcept101} {code} and \href{https://www.cs.cmu.edu/~custom-diffusion/dataset.html} {website}.

\begin{table}[!h]
\centering
\setlength{\tabcolsep}{5pt}
\resizebox{\linewidth}{!}{
\begin{tabular}{ll c c  }
\toprule
& \textbf{Method}
& \shortstack[c]{\textbf{Text-alignment} } 
& \shortstack[c]{\textbf{Image-alignment} } \\
\midrule
\multirow{4}{*}{\shortstack[c]{\textbf{Single-}\\ \textbf{Concept} } }  & Textual Inversion    & 0.612 & \textbf{0.752}  \\
& DreamBooth & 0.752   & \textbf{0.752} \\
& Ours    & \textbf{0.760} & 0.744 \\
\midrule

\multirow{3}{*}{\shortstack[c]{\textbf{Multi-}\\ \textbf{Concept} } } 
& DreamBooth & 0.738 & 0.662  \\
& Ours (Optimization)& 0.763 & 0.658 \\
& Ours (Joint)    & \textbf{0.757} & \textbf{0.668} \\ 
\bottomrule
\vspace{-10pt}
\end{tabular}
}
\vspace{-8pt}
\caption{\textbf{Quantitative comparison on CustomConcept101.} On single-concepts, our method is better on text-alignment and slightly worse on image-ailgnment compared to DreamBooth. For multi-concept, our method outperforms DreamBooth on average. The metrics are calculated over $100$ images generated using $20$ prompts for single-concept and $120$ images generated using $12$ prompts for multi-concept and then averaged across all models. We used DDPM sampler with $200$ steps in each case.}

\label{tbl:results_customconcept101}
\vspace{-6pt}
\end{table}

\section{Multi-Concept Optimization Based Method}\lblsec{derivation}
To compose multiple concepts together, we solved a constrained least squares problem (introduced in \refsec{multi-concept}, \refeq{formulation}). We solved the problem in closed form as shown in \refeq{solution} in the main paper. Here, we show the full derivation. We first restate the objective below.
\begin{equation}
    \begin{split}
     \hat{W} = \argmin_{W} ||  W C_{\textnormal{reg}}^\top - & W_{0}C_{\textnormal{reg}}^\top || \\
     \textnormal{s.t. } \hspace{1mm}  WC^\top = V, \textnormal{ whe}& \textnormal{re } C = [\c_1 \cdots \c_N]^\top \\
     \textnormal{and } & V = [W_{1}\c_1^\top \cdots W_{N}\c_N^\top]^\top.
    \end{split}\lbleq{supp_objective}
\end{equation}
\noindent Here, the matrix norms are the Frobenius norm, $W_0 \in \mathbb{R}^{o \times d}$ is the matrix from the pretrained model, $C \in \mathbb{R}^{s \times d}$ is the text features of dimension $d$. These are compiled of $s$ target words across all $N$ concepts, with all captions for each concept flattened out and concatenated. Similarly, $C_{\textnormal{reg}} \in \mathbb{R}^{s_{\textnormal{reg}} \times d} $ consists of text features of $\smallsim1000$ randomly sampled captions for regularization. 

By using the method of Lagrange multipliers~\cite{boyd2004convex}, we need to minimize the following objective:
\begin{equation}
    \begin{split}
       L = \frac{1}{2} || W C_{\textnormal{reg}}^\top - W_{0}C_{\textnormal{reg}}^\top || -  \textnormal{trace}(\v( WC^\top - V)),
    \end{split}
\end{equation}
\noindent here $\v \in \mathbb{R}^{s \times o} $ is the Lagrangian multiplier corresponding to the constraints. Differentiating the above objective and equating it to $0$, we obtain:
\begin{equation}
    \begin{split}
       W C_{\textnormal{reg}}^\top C_{\textnormal{reg}} - W_{0}C_{\textnormal{reg}}^\top C_{\textnormal{reg}} - \v^\top C = 0 \\
      \implies W = W_0 + \v^\top C(C_{\textnormal{reg}}^\top C_{\textnormal{reg}})^{-1}.
    \end{split}
\end{equation}

\noindent We assume $C_{\textnormal{reg}}$ is non-degenerate. Using the above solution in \refeq{supp_objective}, $ WC^\top = V$, we obtain:
\begin{equation}
    \begin{split}
      & (W_0 + \v^\top C(C_{\textnormal{reg}}^\top C_{\textnormal{reg}})^{-1})C^\top = V 
     \\
     & \textnormal{Let } \d = C(C_{\textnormal{reg}}^\top C_{\textnormal{reg}})^{-1} \\
     & \v^\top = ( V -  W_{0}C^\top )( \d C^\top )^{-1}.
    \end{split}
\end{equation}

\begin{figure}[!t]
    \centering
    \includegraphics[width=0.48\linewidth]{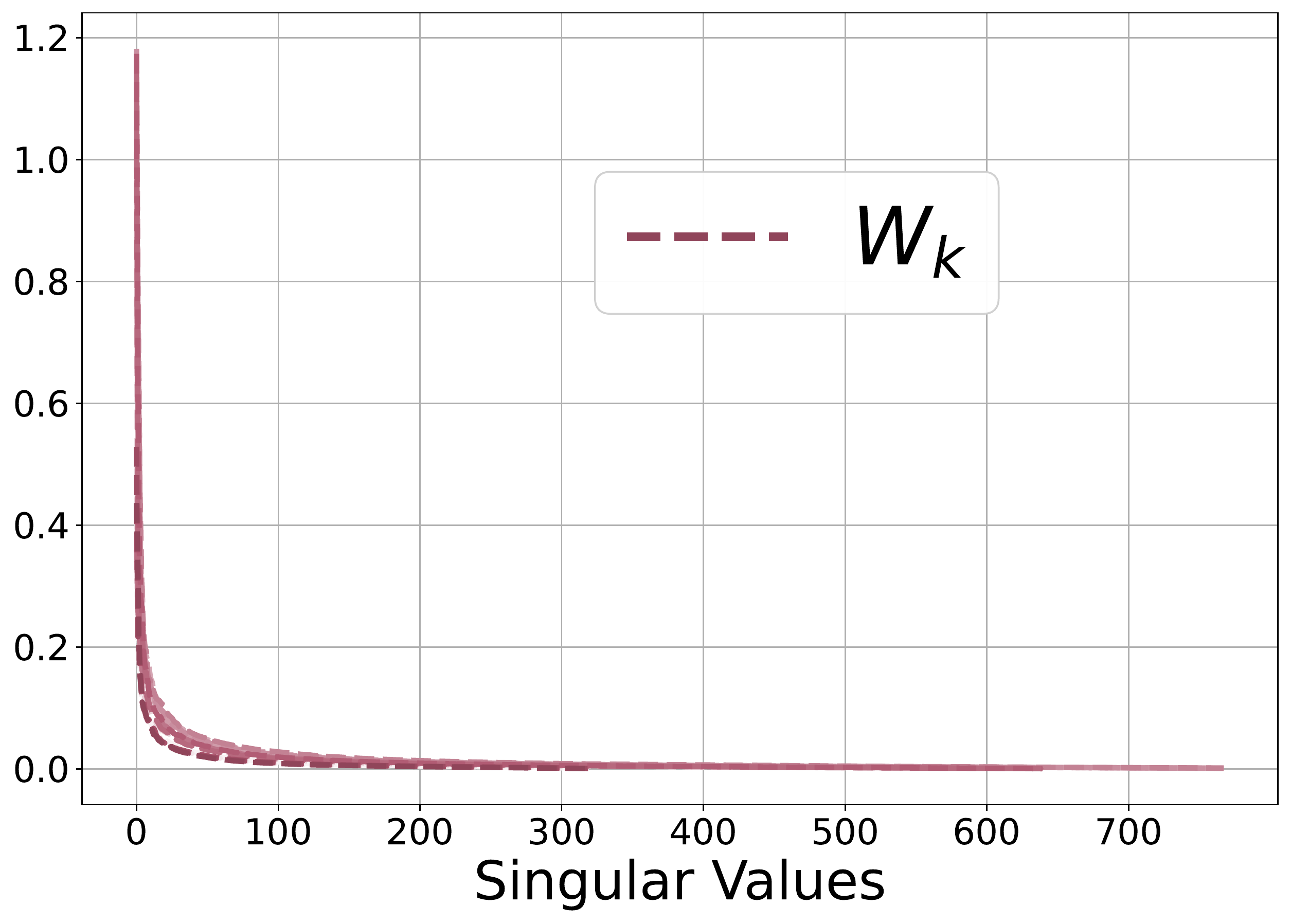}
    \includegraphics[width=0.48\linewidth]{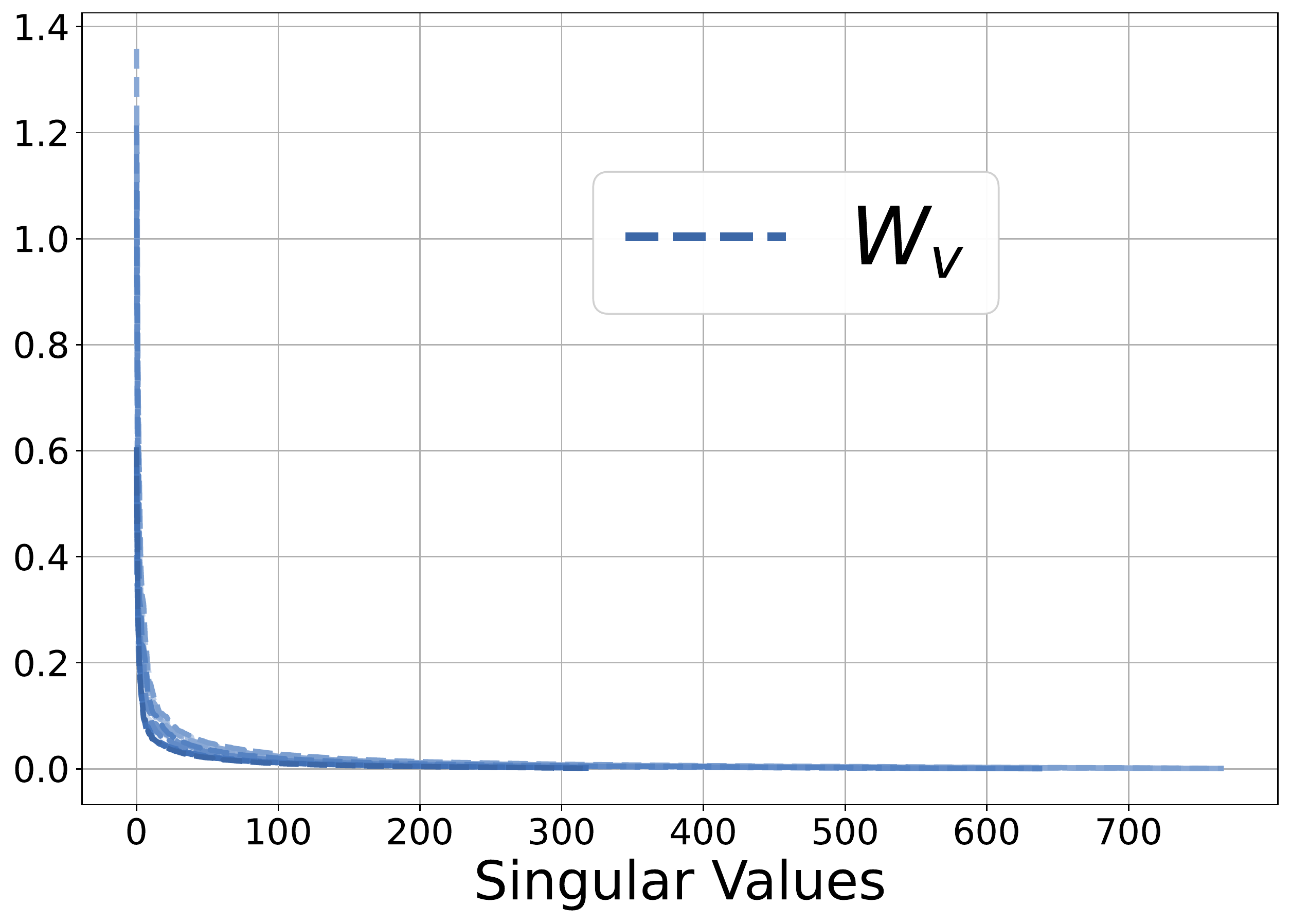}
    \vspace{-10pt}
    \caption{{\textbf{Analysis of singular values} of the difference of the key and value projection matrices in the cross-attention layers between pretrained and fine-tuned model. As shown in the plot, the singular values drop to 0, suggesting that we can approximate the difference matrix with a low-rank matrix.
    }}
    \lblfig{singular_value}
    \vspace{-10pt}
\end{figure}

\begin{figure}[!t]
    \centering
    \includegraphics[width=\linewidth]{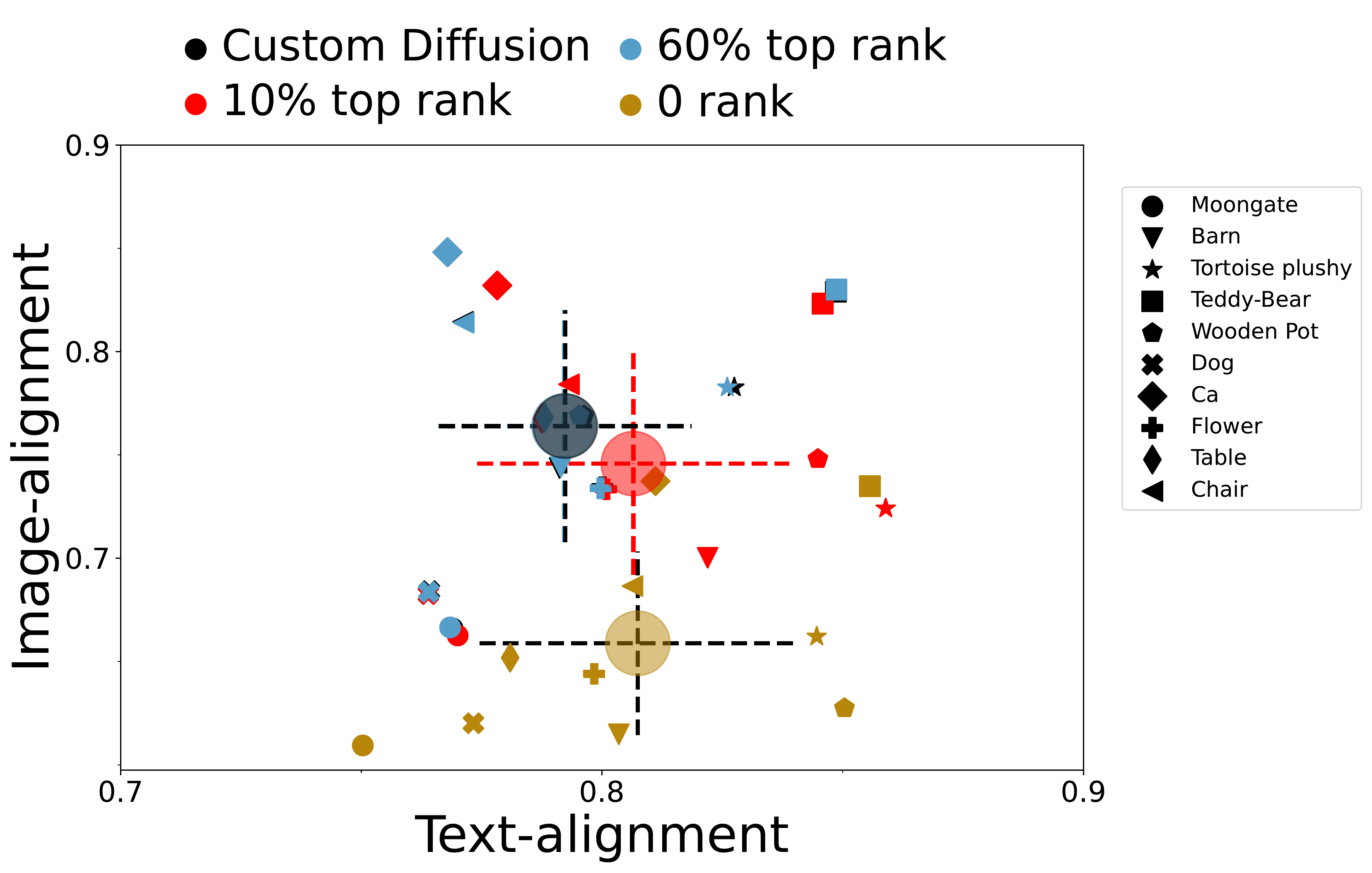}
    \vspace{-10pt}
    \caption{{\textbf{Quantitative results with compressed models}. We save the low-rank approximation of the difference between the pretrained and fine-tuned model updated weights. Even with the top $60\%$ rank ($5\times$ compression), the performance remains similar (overlapping blue and black points). As we increase the compression, the image-alignment score decreases, and the model approaches the pretrained model weights with high text-alignment, as  illustrated with qualitative samples in \reffig{supp_compression}. 
    }}
    \lblfig{compression_num}
    \vspace{-15pt}
\end{figure}

\begin{figure}[!t]
    \centering
    \includegraphics[width=\linewidth]{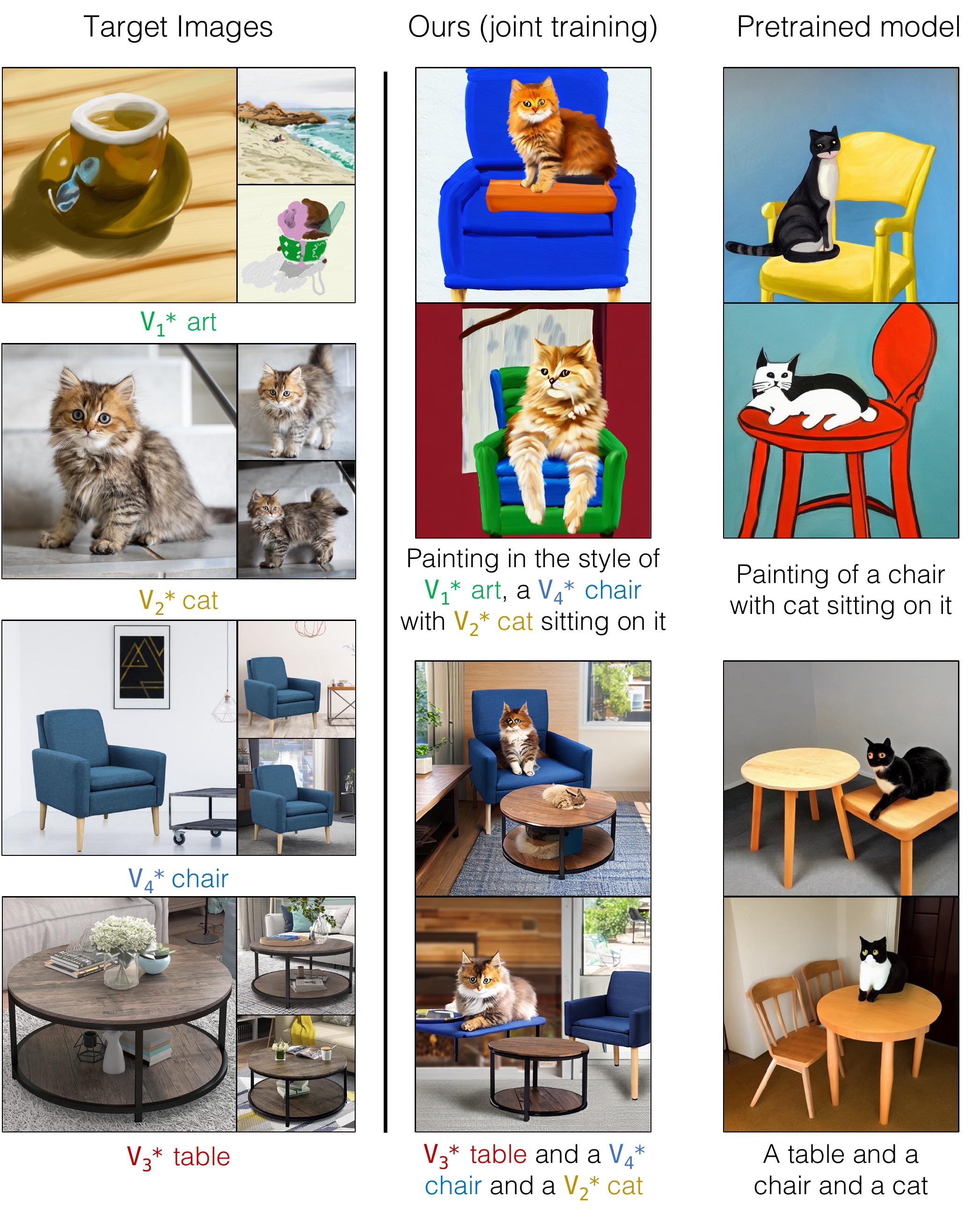}
    \vspace{-25pt}
    \caption{{\textbf{Three-concept fine-tuning results.} We can also compose two objects with a style or three new objects in the same scene for some combination of concepts.  
    }}
    \label{fig:multi_concept_limitation}
    \vspace{-12pt}
\end{figure}

\section{Experiments}\lblsec{exprmore}

\begin{figure*}[!t]
    \centering
    \includegraphics[width=\linewidth]{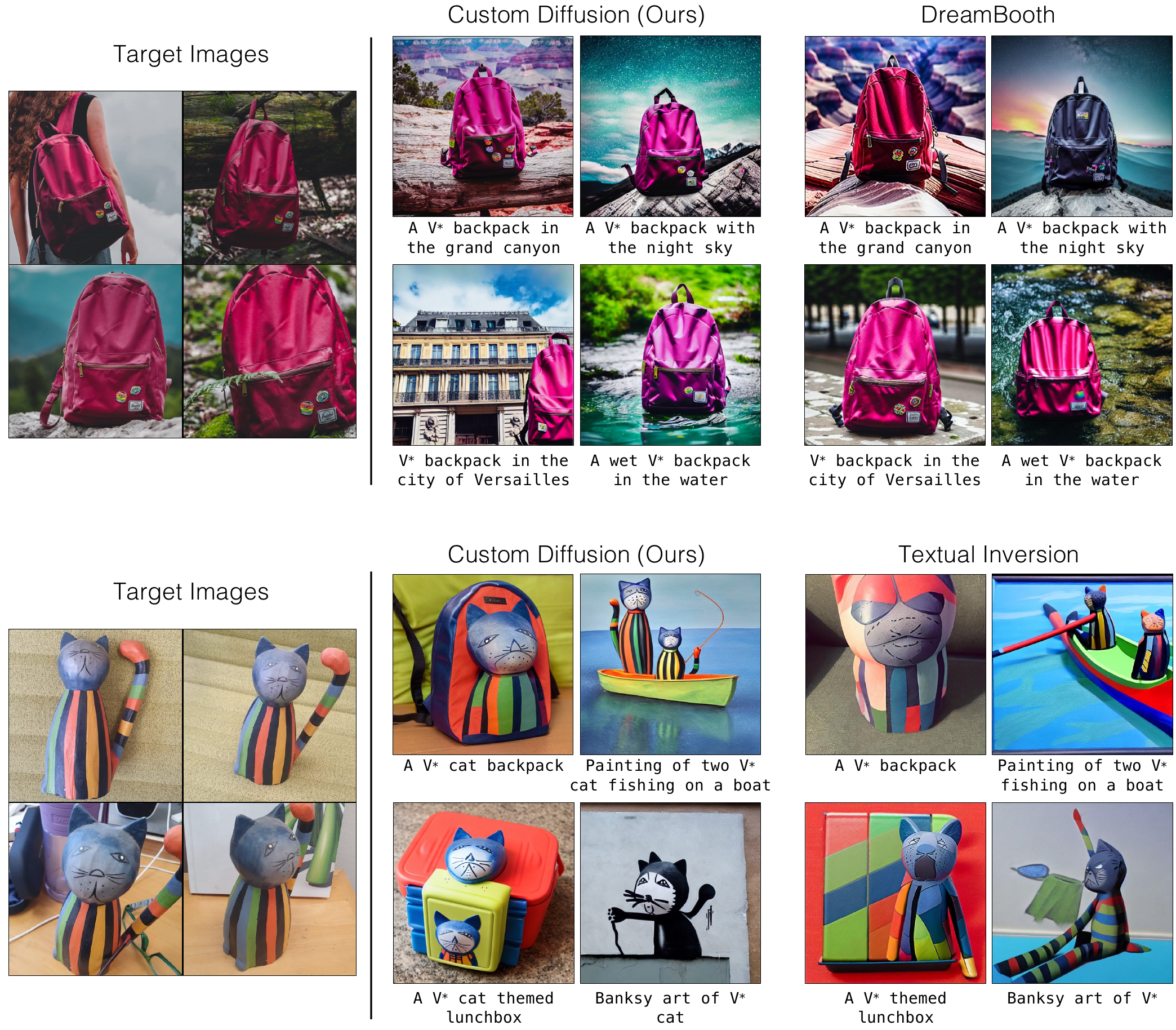}
    \vspace{-10pt}
    \caption{{\textbf{Results of our method on DreamBooth~\cite{ruiz2022dreambooth} and Textual Inversion~\cite{gal2022image} datasets}. Our method works similarly or better in some instances, for example, {\menlo V$^*$ backpack with the night sky}, and for {\menlo a V$^*$ cat backpack}. The samples are generated with $200$ steps of the DDPM sampler, scale $6$.
    }}
    \label{fig:supp_dreambooth_ti}
    \vspace{-10pt}
\end{figure*}

In this section, we show more experiments, including ablation and analysis of the limitations of our method.

\myparagraph{Comparison with DreamBooth and Textual Inversion.} We also show a comparison of our method with DreamBooth and Textual Inversion on the target concept images from the respective works. \reffig{supp_dreambooth_ti} shows the sample generations for the same input prompts. Our method performs better than baselines in text-alignment while maintaining the visual similarity to target images.

\begin{figure*}[!t]
    \centering
    \includegraphics[width=\linewidth]{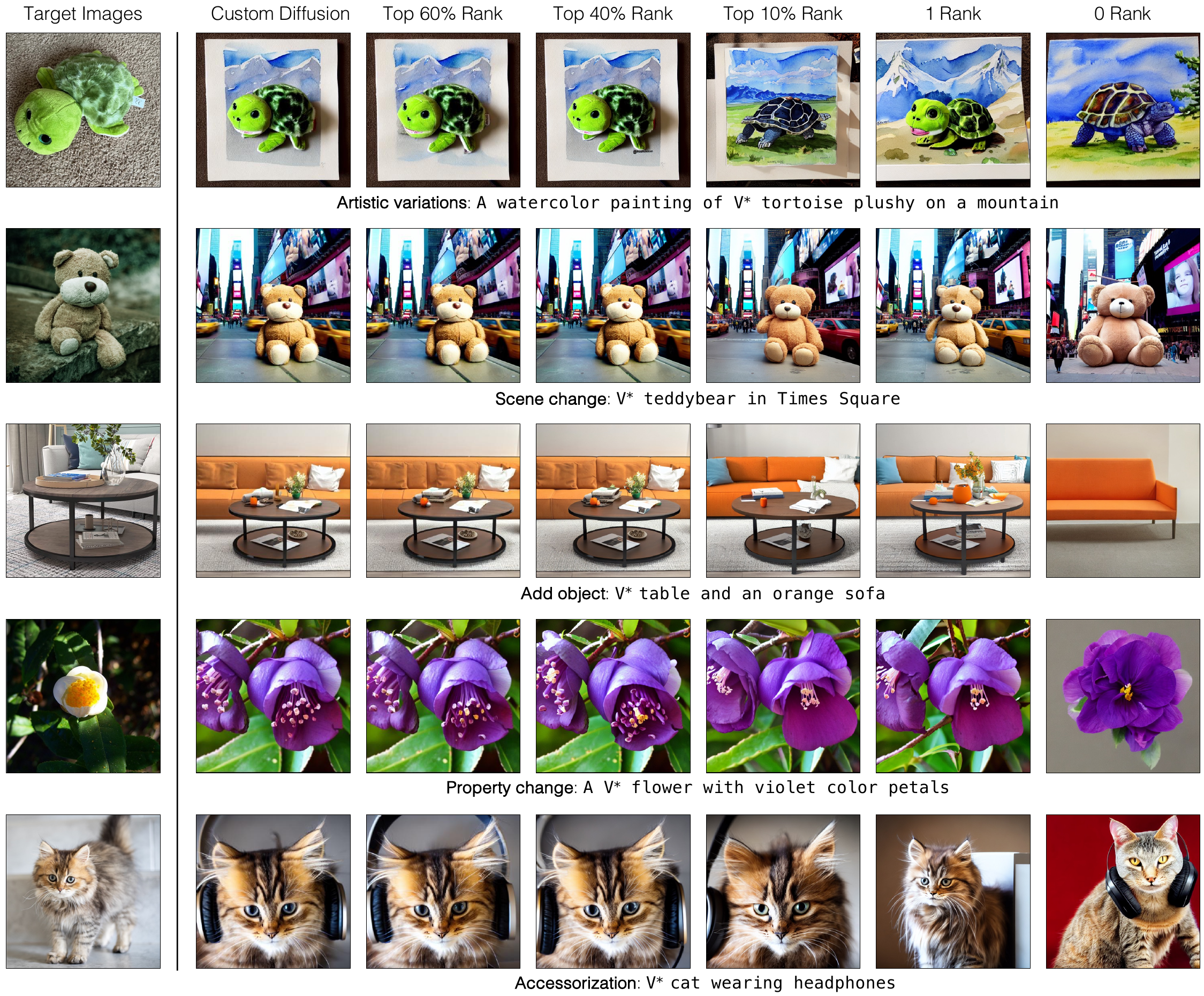}
    \vspace{-10pt}
    \caption{{\textbf{Qualitative results on fine-tuned model's compression for reduced storage requirements.} We save the low-rank approximation of the difference between the pretrained model and fine-tuned model updated weights. The storage requirements of models from left to right are 75MB, 15MB, 5MB, 1MB,  0.1MB, and 0.08MB (to save the optimized {\menlo V$^*$}). Even with $5\times$ compression with top $60\%$ singular values, the performance remains similar. Top k$\%$ implies singular values till the rank where cumulative sum is k$\%$ of total sum of singular values. As we increase the compression, the image-alignment score decreases, as evident from sample generations not being similar to target images, especially in the case of tortoise plushy, teddybear, and cat.
    }}
    \label{fig:supp_compression}
\end{figure*}

\myparagraph{Model compression.}
We analyze the singular values of the difference between updated key and value projection matrices and the corresponding pretrained matrices in all cross-attention layers of the model. As shown in \reffig{singular_value}, the singular values steeply drop. This suggests that the difference matrix can be approximated well with a low-rank decomposition. Thus, we perform SVD and only save the low-rank factorization of the difference matrices. This can reduce the memory storage for each concept further. \reffig{compression_num} and \ref{fig:supp_compression} shows the quantitative and qualitative results with decreasing compression ratio. Even with an approximation using only the top $60\%$ of singular values ($5 \times$ compression in model storage), the performance is similar. But as we increase the compression to only include the top $1\%$ of singular values, the image-alignment decreases. Top k$\%$ implies singular values till the rank where cumulative sum is k$\%$ of total sum of singular values. We also attempted to enforce a low-rank update to the key, value matrices during fine-tuning but observed the results to be sub-optimal~\cite{frankle2018lottery}. More sample generations are shown on our \href{https://www.cs.cmu.edu/~custom-diffusion/}{website}.

\begin{figure*}[!t]
    \centering
    \includegraphics[width=\linewidth]{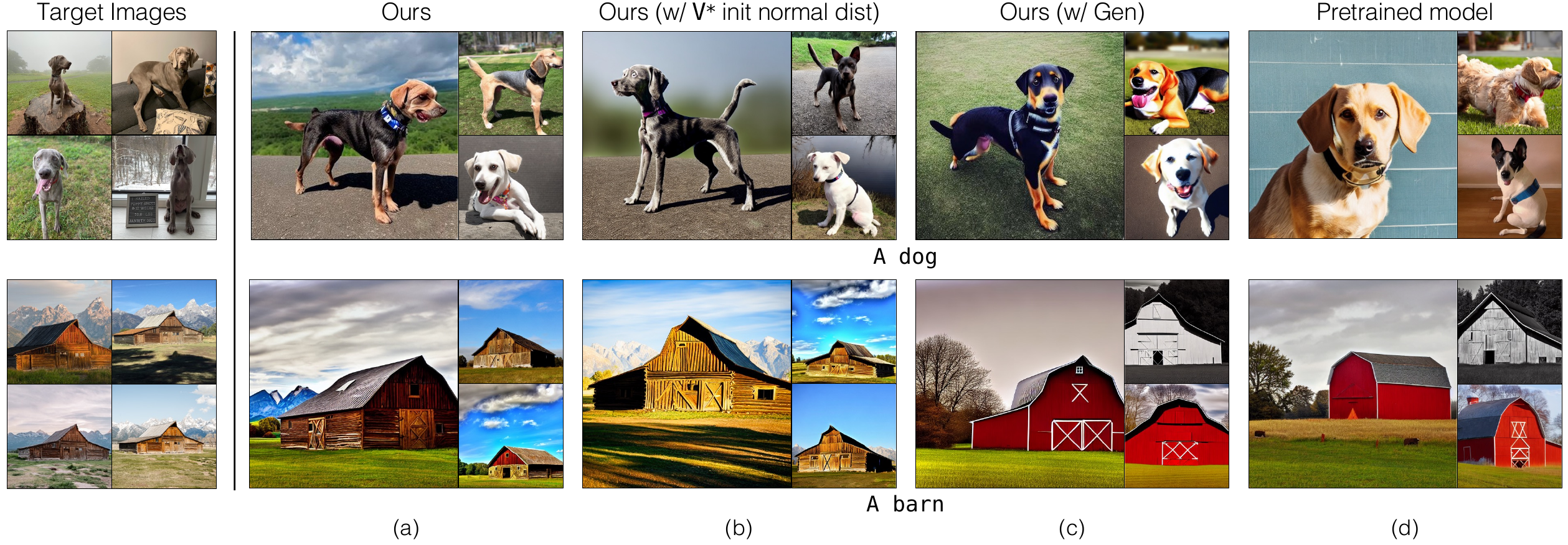}
    \vspace{-20pt}
    \caption{{\textbf{Qualitative analysis of Choice of {\menlo V$^*$} (Ours w/ random init {\menlo V$^*$}) and generated images as regularization (Ours w/ Gen).} We show the generated samples for the original category word, for e.g., images generated on the prompt {\menlo a dog} for models fine-tuned on {\menlo V$^*$ dog}. We also show the sample generated by pretrained model in column (d) for comparison. Column (b) shows that initializing {\menlo V$^*$} randomly from the normal distribution of existing token embeddings and then optimizing leads to more shifts in original category words getting mapped to target images compared to our method in column (a). %
    Similarly, with generated images as regularization, the quality of samples gets worse for the original category as shown in column (c).
    }}
    \lblfig{choice_v}
\end{figure*}

\myparagraph{Choice of {\menlo V$^*$}.} In all experiments, we initialized the unique modifier token with the token-id $42170$. During joint training with two concepts, we initialize the other with token-id $47629$ in the pretrained CLIP tokenizer used in Stable Diffusion. We ablate our choice of {\menlo V$^*$} with -- (1) Random initialization with the mean and standard deviation of existing token embeddings and then optimizing the modifier token, (2) not optimizing the modifier token, once initialized with the rarely occurring token. The quantitative results are shown in \reftbl{choice_v}. We observe that not optimizing {\menlo V$^*$} leads to significantly lower image-alignment. Compared to random initialization of {\menlo V$^*$}, our method has higher image-alignment and lower text-alignment. But we observe that the generated samples with the prompt { \menlo a \{category\}} shift more towards target image distribution of {\menlo V$^*$ category}, compared to our final method, as shown in \reffig{choice_v}.

\myparagraph{Three concept fine-tuning.} We test the limit of our multi-concept method and show results on three concepts in \reffig{multi_concept_limitation}. Our method is able to compose two new objects with a new style or synthesize three new objects.

\begin{figure*}[!t]
    \centering
    \includegraphics[width=\linewidth]{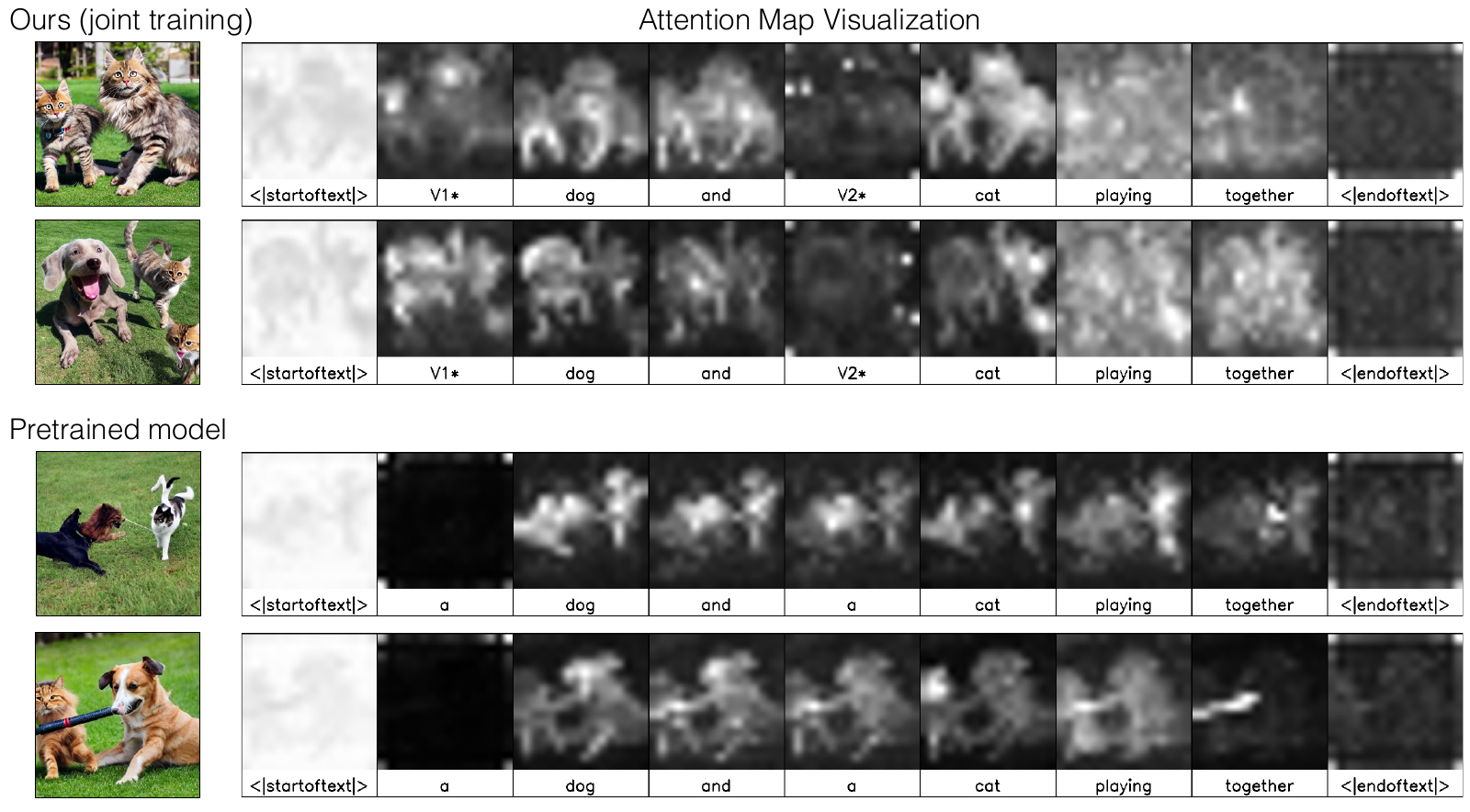}
    \vspace{-20pt}
    \caption{{\textbf{Attention map visualization of failed compositions.} We show the average attention across timestep and layers for each word (token). For both our and pretrained models, the attention map of ``cat'' and ``dog'' overlap more often. This can be one of the reasons for the failed compositions. 
    }}
    \label{fig:supp_limitation_attn}
    \vspace{-2pt}
\end{figure*}

\myparagraph{Limitations of our multi-concept fine-tuning.}
As shown in \reffig{results_multi_concept_limitation} in the paper, our method fails at difficult compositions like generating personal cat and dog in the same scene. We observe that the pretrained model also struggles with such compositions and hypothesize that our model inherits these limitations. Here, we analyze the attention map of each token on the latent image features in \reffig{supp_limitation_attn}. The ``dog'' and ``cat'' token attention maps are largely overlapping for both our and pretrained models, which might lead to worse composition.

\myparagraph{Generated images as regularization (Ours w/ Gen).}
We showed in \refsec{ablation} (\reftbl{ablation}) that using generated images as regularization leads to similar performance on the target concept but higher KID~\cite{binkowski2018demystifying} on the validation set. We show here that it also leads to artifacts in the generations with category word used to generate images which are used as the regularization set. For example, fine-tuned model on {\menlo V$^*$} dog generates images with saturation artifact for the prompt {\menlo a dog} as shown in \reffig{choice_v}. Since we use a high learning rate compared to DreamBooth (which also uses generated images as regularization). A lower learning rate might mitigate this issue but at the cost of increased training time.

\begin{table*}[!t]
\centering
\setlength{\tabcolsep}{5pt}
\resizebox{\linewidth}{!}{
\begin{tabular}{ll ccccccccc c }
\toprule

& \multirow{2}{*}{\textbf{Methods}}
& \multirow{2}{*}{Barn (7)  }
& \multirow{2}{*}{\shortstack[c]{Tortoise \\ plushy (12)}}
& \multirow{2}{*}{\shortstack[c]{Teddy-\\Bear (7)}}
& \multirow{2}{*}{\shortstack[c]{Wooden\\Pot (4)}}
& \multirow{2}{*}{Dog (10)  }
& \multirow{2}{*}{Cat (5) }
& \multirow{2}{*}{Flower (10) }
& \multirow{2}{*}{Table (4)}
& \multirow{2}{*}{Chair (4)}
& \multirow{2}{*}{\textbf{Mean}}
\\ \\

\midrule

\multirow{3}{*}{\textbf{Text-alignment}} & Ours   & 0.791 & \textbf{0.827} & 0.849 & 0.796 & 0.764 & 0.768 & 0.800   & 0.788 & 0.771 & 0.795  \\

\cdashline{2-12}

& Ours (w/ {\menlo V$^*$} init normal dist.) & 0.817 & 0.809 & 0.853 & 0.807 & \textbf{0.781} & \textbf{ 0.805} & 0.820  & \textbf{0.793} & 0.766 & 0.805 \\

& Ours (w/o {\menlo V$^*$} opt)   & \textbf{0.847} & 0.824 & \textbf{0.864} & \textbf{0.830}  & 0.769 & 0.801 & \textbf{0.823} & 0.787 & \textbf{0.807} &  \textbf{0.816} \\

\midrule

\multirow{3}{*}{\textbf{Image-alignment}} & Ours  & 0.744 & \textbf{0.783} & \textbf{0.829} & 0.769 & \textbf{0.684} & \textbf{0.848} & \textbf{0.734} & 0.768 & \textbf{0.814} & \textbf{0.774} \\

\cdashline{2-12}

& Ours (w/ {\menlo V$^*$} init normal dist.) & \textbf{0.747} & 0.780  & 0.829 & \textbf{0.787} & 0.670  & 0.785 & 0.709 & \textbf{0.772} & 0.811 & 0.765 \\

& Ours (w/o {\menlo V$^*$} opt)   & 0.730  & 0.755 & 0.784 & 0.786 & 0.663 & 0.757 & 0.683 & 0.688 & 0.757 & 0.733  \\

\midrule

\multirow{3}{*}{\shortstack[c]{ \textbf{KID}  ($\times 10^3$) \\ (Validation)  }}

& Ours  &  \textbf{09.00} & 26.82 & \textbf{40.33} &  \textbf{08.77} & 19.46  & 27.39 & 36.47 & \textbf{15.77} & \textbf{17.94} & 22.43 \\  

\cdashline{2-12}

& Ours (w/ {\menlo V$^*$} init normal dist.) &  09.99 & \textbf{21.76} & 44.68 & 12.27 & \textbf{15.09} & \textbf{25.26}  & 32.97 & 15.26 & 21.28 & 22.06 \\

& Ours (w/o {\menlo V$^*$} opt)  & 10.22 & 23.75 & 41.64 & 11.97 & 18.19 & 26.41 & \textbf{28.83} & 16.56 & 19.20 & \textbf{21.86} \\

\bottomrule
\end{tabular}}
\vspace{-8pt}
\caption{\textbf{Quantitative results of different choices for modifier token {\menlo V$^*$}}. We ablate our method with two settings -- (1) \textbf{Ours (w/ {\menlo V$^*$} init normal dist.)}, where we initialize the modifier token randomly with mean and standard deviation of the existing token embeddings, and then optimize during training. (2) \textbf{Ours (w/o {\menlo V$^*$} opt)}, i.e., not optimizing the token once initialized with the rare occurring token. We observe that not optimizing  {\menlo V$^*$} leads to worse results on image-alignment, i.e., the model is not able to learn the target concept. Similarly, random initialization with the normal distribution also results in lower image-alignment and higher text-alignment, but as shown in \reffig{choice_v}, the category word mapping shifts to target images. %
}

\label{tbl:choice_v}
\vspace{-5pt}
\end{table*}

\begin{table*}[!t]
\centering
\resizebox{\linewidth}{!}{
\begin{tabular}{ll cccccccccc  }
\toprule

& \multirow{2}{*}{\textbf{Methods}}
& \multirow{2}{*}{\shortstack[c]{Moongate\\ (135)~\cite{liu2020towards}}}
& \multirow{2}{*}{Barn (7)  }
& \multirow{2}{*}{\shortstack[c]{Tortoise \\ plushy (12)}}
& \multirow{2}{*}{\shortstack[c]{Teddy-\\Bear (7)}}
& \multirow{2}{*}{\shortstack[c]{Wooden\\Pot (4)}}
& \multirow{2}{*}{Dog (10)  }
& \multirow{2}{*}{Cat (5) }
& \multirow{2}{*}{Flower (10) }
& \multirow{2}{*}{Table (4)}
& \multirow{2}{*}{Chair (4)}
\\ \\

\midrule

\multirow{2}{*}{MS-COCO FID} & \textbf{Ours} & \textbf{16.05} & \textbf{17.21} & \textbf{16.27} & \textbf{16.71} & \textbf{16.70} & \textbf{16.26} & \textbf{16.95} & \textbf{16.71} & \textbf{16.25} & \textbf{16.99}  \\

& \textbf{DreamBooth}   & 17.35 & 20.36 & 19.61 & 19.45 & 20.01 & 19.10 & 20.57 & 18.57 & 19.39 & 19.35 \\
 
\bottomrule
\end{tabular}}
\vspace{-8pt}
\caption{\textbf{MS-COCO FID evaluation} with fine-tuned models is a standard evaluation metric for text-to-image models. The pretrained stable diffusion model has $16.35$ FID with the same setting of $50$ DDPM sampling steps, scale $6$. Our method for most datasets has a similar FID, which shows that the fine-tuned models are similar to the pretrained model on other unrelated concepts. Since Textual Inversion does not update the model, it has the same FID as the pretrained model.}\label{tbl:mscoco}
\vspace{-5pt}
\end{table*}

\begin{figure*}[!t]
    \centering
    \includegraphics[width=\linewidth]{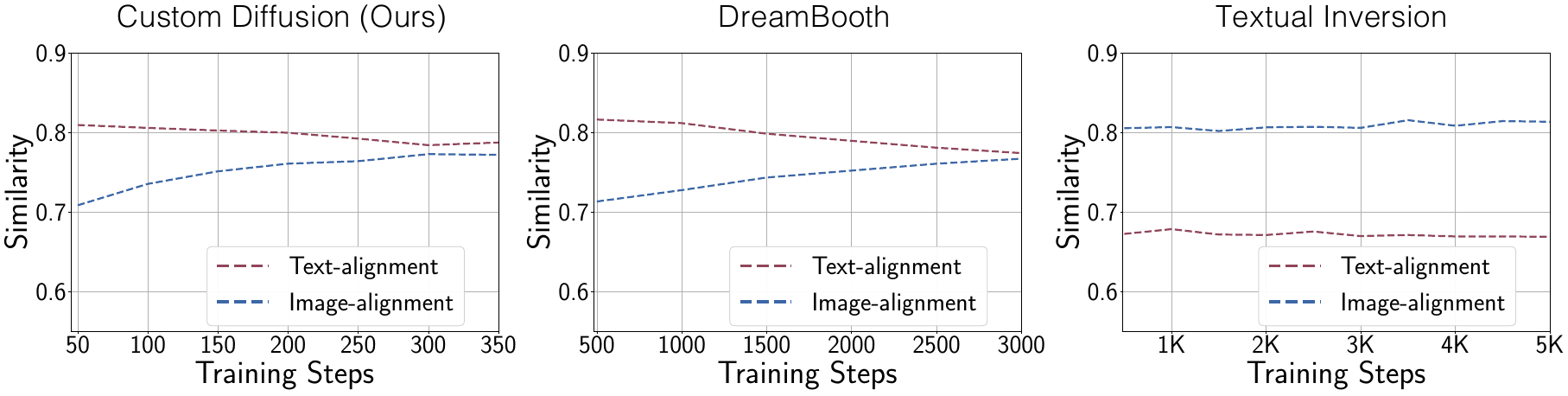}
    \vspace{-10pt}
    \caption{{\textbf{Text- and image-alignment scores} with training steps (mean across datasets). The text-alignment score gradually decreases as we fine-tune the model on the target images. In the case of Textual Inversion, we observe that the new token embedding introduced in the method results in high image-alignment but the text-alignment remains significantly lower across training iterations. Compared to DreamBooth, our method at convergence has higher image text- and image-alignment.
    }}
    \lblfig{training_iter}
    \vspace{-5pt}
\end{figure*}

\begin{table*}[!t]
\centering
\setlength{\tabcolsep}{5pt}
\resizebox{\linewidth}{!}{
\begin{tabular}{ll cccccccccc c }
\toprule

& \multirow{2}{*}{\textbf{Methods}}
& \multirow{2}{*}{\shortstack[c]{Moongate\\ (135)~\cite{liu2020towards}}}
& \multirow{2}{*}{Barn (7)  }
& \multirow{2}{*}{\shortstack[c]{Tortoise \\ plushy (12)}}
& \multirow{2}{*}{\shortstack[c]{Teddy-\\Bear (7)}}
& \multirow{2}{*}{\shortstack[c]{Wooden\\Pot (4)}}
& \multirow{2}{*}{Dog (10)  }
& \multirow{2}{*}{Cat (5) }
& \multirow{2}{*}{Flower (10) }
& \multirow{2}{*}{Table (4)}
& \multirow{2}{*}{Chair (4)}
& \multirow{2}{*}{\textbf{Mean}}
\\ \\

\midrule

\multirow{4}{*}{\shortstack[c]{ Text-alignment}} &

  \textbf{Textual Inversion}       & 0.658 & 0.739 & 0.646 & 0.713 & 0.726 & 0.613 & 0.643 & 0.648 & 0.651 & 0.658 & 0.670 \\
& \textbf{DreamBooth}             & \textbf{0.778} & \textbf{0.839} & 0.789 & 0.850  & 0.791 & \textbf{0.764} & 0.789 & \textbf{0.824} & 0.715 & 0.676 & 0.781 \\
\cdashline{2-13}

& \textbf{Ours (w/ fine-tune all)} & 0.762 & 0.782 & 0.822 & \textbf{0.854} & \textbf{0.820}  & 0.752 & \textbf{0.792} & 0.803 & 0.763 & 0.767 & \textbf{0.795} \\
& \textbf{Ours}     & 0.772 & 0.791 & \textbf{0.828} & 0.848 & 0.798 & \textbf{0.764} & 0.768 & 0.800   & \textbf{0.787} & \textbf{0.771} & \textbf{0.795} \\

\midrule 
\multirow{4}{*}{\shortstack[c]{ Image-alignment}} &

\textbf{Textual Inversion}       & \textbf{0.740}  & \textbf{0.753} & 0.755 & \textbf{0.885} & \textbf{0.802} & \textbf{0.760}  & \textbf{0.909} & \textbf{0.815} & \textbf{0.891} & \textbf{0.872} & \textbf{0.827} \\
 & \textbf{DreamBooth}            & 0.608 & 0.679 & \textbf{0.847} & 0.822 & 0.788 & 0.674 & 0.786 & 0.688 & 0.846 & 0.863 & 0.776 \\

 \cdashline{2-13}
&  \textbf{Ours (w/ fine-tune all)} & 0.662 & 0.699 & 0.822 & 0.804 & 0.767 & 0.669 & 0.810  & 0.670  & 0.725 & 0.772 & 0.748 \\
&  \textbf{Ours}                  & 0.665 & 0.744 & 0.784 & 0.829 & 0.769 & 0.684 & 0.849 & 0.735 & 0.767 & 0.814 & 0.775 \\

\midrule

\multirow{5}{*}{\shortstack[c]{ KID  ($\times 10^3$) \\ (Validation)  }}

& \textbf{Pretrained Model}  & 11.33 & 11.18 & 33.51 & 39.82 & 12.88 & 25.40 & 35.01 & 30.96 & \textbf{13.35} &  \textbf{9.26} & 22.27   \\   

\cdashline{2-13}

& \textbf{Textual Inversion}     & 11.33 & 11.18 & 33.51 & 39.82 & 12.88 & 25.40 & 35.01 & 30.96 & 13.35 &  9.26 & 22.27   \\
& \textbf{DreamBooth}    & 20.80 & 12.975 & 36.45 & 42.93 & 15.80 & 44.14 & 67.41 & 37.55 & 21.07 & 26.16 & 32.53  \\ 

\cdashline{2-13}

& \textbf{Ours (w/ fine-tune all)} &  9.04 &  9.35 & 28.55 & 42.78 & 10.60 & 19.58 & \textbf{14.79} & \textbf{22.95} & 18.69 & 16.34 & \textbf{19.27}  \\

& \textbf{Ours} &  \textbf{7.65} &  \textbf{9.00} & \textbf{26.82} & \textbf{40.33} &  \textbf{8.77} & \textbf{19.46}  & 27.39 & 36.47 & 15.77 & 17.94 & 20.96 \\  

\bottomrule
\end{tabular}}
\vspace{-8pt}
\caption{\textbf{Quantitative evaluation on single-concept fine-tuning}. \textit{\underline{First and Second row}}: we show the text- and image-alignment in CLIP feature space (higher is better for both). All metrics are calculated with $1$K generated samples across $20$ prompts for each dataset. Our method performs better than baselines when averaged across datasets. We show the trend with training steps in \reffig{training_iter}. \textit{\underline{Bottom row}}: KID between real validation set images ($500$) and generated images ($1$K) with the same caption. Since our method uses a regularization set of real images, it achieves lower KID than baselines and even improves slightly over the pretrained model except on ``Table'' and ``Chair''. Textual Inversion has the same KID as the pretrained model since the diffusion model is not updated in the method. We also evaluate our models with FID~\cite{fid} on MS-COCO~\cite{lin2014microsoft} in \reftbl{mscoco} in the Appendix. We use the DDPM sampler with $50$ steps and scale $6$ for both metrics. The number of training images is shown in brackets.
}

\label{tbl:ind_values_single_concept}
\vspace{-8pt}
\end{table*}

\section{Evaluation}\lblsec{evalmore}

\myparagraph{Text- and image-alignment scores with training iterations.}
There is usually a trade-off between text-alignment and image-alignment. High image-alignment leads to a decrease in text-alignment where sample generations have less variance and are close to input target images. We show the trend of text- and image-alignment in \reffig{training_iter}, which shows that initially the model has high text-alignment (as pretrained model) but low image-alignment and with training the curves for text-/image-alignment gradually gets worse/better. \reftbl{ind_values_single_concept} and \ref{tbl:supp_multi-concept-training_eval} shows the individual text- and image-alignment scores for the single-concept and multi-concept fine-tuning. To measure text-alignment, we remove the modifier token {\menlo V$^*$} from the text prompt for extracting CLIP text features.

\myparagraph{MS-COCO FID evaluation for all models.}
We also evaluate MS-COCO FID for our models and DreamBooth. As shown in \reftbl{mscoco}, our method has lower FID and only slightly worse occasionally compared to the $16.35$ FID of the pretrained model. This suggests our method doesn't change the generated image distribution on unrelated concepts. We measure FID~\cite{fid} using \texttt{clean-fid} library~\cite{cleanfid}.

\section{Implementation and Experiment Details}\lblsec{details}

We describe additional training details for our method and baselines~\cite{gal2022image,ruiz2022dreambooth}.

\myparagraph{Datasets.} All datasets in the paper were captured manually or downloaded from Unsplash except Moongate~\cite{liu2020towards}.

\begin{table*}[!t]
\centering
\setlength{\tabcolsep}{5pt}
\resizebox{\linewidth}{!}{
\begin{tabular}{lll ccccc c }
\toprule
& \multicolumn{2}{l}{\textbf{Methods}}
& \begin{tabular}{@{}c@{}} Moongate + Dog \end{tabular}
& \begin{tabular}{@{}c@{}} Cat + Chair \end{tabular}
& \begin{tabular}{@{}c@{}} Wooden Pot + Cat \end{tabular} %
&\begin{tabular}{@{}c@{}}  Wooden Pot + Flower \end{tabular} %
& \begin{tabular}{@{}c@{}} Table + Chair  \end{tabular} & {\bf Mean} \\
\midrule

\multirow{6}{*}{Text-alignment}
& \multicolumn{2}{l}{\textbf{DreamBooth}}      & 0.767 & 0.783 & 0.774 & 0.860  & 0.732 & 0.783 \\

& \multicolumn{2}{l}{\textbf{Textual Inversion}}   & 0.538 & 0.445 & 0.542 & 0.552 & 0.644 & 0.544 \\

\cdashline{2-9}

& \multirow{4}{*}{\textbf{Ours}} & \textbf{w/ fine-tune all}  & \textbf{0.797} & 0.742 & 0.800   & 0.856 & 0.745 & 0.787 \\

& & \textbf{Sequential}  & 0.785 & 0.736 & \textbf{0.863} & 0.862 & 0.740  & 0.797 \\

& & \textbf{Optimization}  & 0.786 & 0.774 & 0.806 & \textbf{0.870}  & \textbf{0.766} & 0.800 \\

& & \textbf{Joint}     & 0.793 & \textbf{0.798} & 0.833 & 0.845 & 0.737 & \textbf{0.801} \\

\midrule

\multirow{6}{*}{\shortstack[c]{ Image-alignment, \\ (target1)}}
& \multicolumn{2}{l}{\textbf{DreamBooth}}     & 0.492 & 0.715 & \textbf{0.719} & 0.697 & 0.817 & 0.688 \\

& \multicolumn{2}{l}{\textbf{Textual Inversion}}  &  \textbf{0.675} & 0.571 & 0.539 & 0.531 & 0.822 & 0.627 \\

\cdashline{2-9}

& \multirow{4}{*}{\textbf{Ours}} & \textbf{w/ fine-tune all} & 0.584 & 0.736 & 0.594 & 0.699 & 0.817 & 0.686  \\

& & \textbf{Sequential}  & 0.534 & 0.696 & 0.636 & 0.671 & 0.833 & 0.674\\ 

& & \textbf{Optimization}  & 0.583 & \textbf{0.792} & 0.580  & 0.683 & 0.786 & 0.684  \\

& & \textbf{Joint}    & 0.592 & 0.674 & 0.654 & \textbf{0.758} & \textbf{0.837} & \textbf{0.702}  \\

\midrule

\multirow{6}{*}{\shortstack[c]{ Image-alignment, \\ (target2)}}

& \multicolumn{2}{l}{\textbf{DreamBooth}}     & \textbf{0.656} & 0.737 & 0.633 & 0.633 & \textbf{0.839} & 0.699 \\

& \multicolumn{2}{l}{\textbf{Textual Inversion}}   & 0.473 & 0.614 & 0.673 & 0.580  & 0.831 & 0.634 \\

\cdashline{2-9}

& \multirow{4}{*}{\textbf{Ours}} & \textbf{w/ fine-tune all}  & 0.592 & 0.646 & 0.815 & 0.644 & 0.783 & 0.695 \\

& & \textbf{Sequential}  & 0.641 & 0.762 & 0.748 & 0.660  & 0.819 & 0.725 \\

& & \textbf{Optimization}    & 0.598 & 0.639 & \textbf{0.819} & \textbf{0.675} & 0.803 & 0.706 \\

& & \textbf{Joint}     & 0.582 & \textbf{0.767} & 0.757 & 0.640  & 0.807 & \textbf{0.710} \\

\bottomrule
\end{tabular}}
\vspace{-8pt}
\caption{\textbf{Text-alignment and image-alignment with the two target concepts in CLIP feature space on multi-concept fine-tuning.} We evaluate each composition pair on $400$ images generated using $8$ prompts with $200$ steps of DDPM sampler and scale=$6$. A high score on one metric at the cost of worse performance on other metrics leads to overall worse results, as our qualitative samples show as well. Our method performs better than concurrent methods on the average metric in all settings except Table+Chair, where most methods perform comparably. 
}

\label{tbl:supp_multi-concept-training_eval}
\end{table*}

\myparagraph{Custom Diffusion (ours).}
As mentioned in \refsec{multi-concept}, we train with a batch size of $8$ and learning rate $10^{-5}$, which is scaled by batch size for an effective learning rate of $8 \times 10^{-5}$. We train for $250$ steps for single-concept experiments and $500$ steps for multi-concept. During training, we randomly resize the target images to $1.2-1.4\times$ every 1 out of 3 times and append {\menlo zoomed in} or {\menlo close up} to the text prompt. The rest of the time the target image is randomly resized to $0.4-1.0\times$ and if the resize ratio is less than $0.6$ we append {\menlo far away} or {\menlo very small} to the text prompt, and only propagate the loss in the valid image region. Since fine-tuning is done only for a few iterations, we do not notice any augmentation leaking in the fine-tuned model. We also detach the start token embedding during fine-tuning. For selecting the rare token as the modifier token {\menlo V$^*$}, we count the occurrence of the total $49408$ tokens in $200K$ captions sampled from the LAION-400M dataset. We then select the token with $\smallsim 5-10$ occurrences, with alphabetic representation, and not a substring of another token. During training, we oversample the target images to keep the ratio of regularization samples ($200$) and target samples the same. 

\myparagraph{Ours (w/ fine-tune all).}
When fine-tuning all parameters, we reduce the learning rate to $8 \times 10^{-6}$, which works better than $8 \times 10^{-5}$ and train for $500$ steps with a batch size of $8$. For multi-concept, we train for $1000$ iterations. The rest of the settings are the same as our final method.

\myparagraph{Textual Inversion~\cite{gal2022image}}
We train with the recommended batch size of $8$, a learning rate of $0.005$ (scaled by batch size for an effective learning rate of $0.04$) for $5000$ steps. The new token is initialized with the category word, e.g., ``dog''. In cases when the category word is represented by multiple tokens, e.g., ``tortoise plushy'', we use a single word approximation like ``plush'', similarly ``gate'' for ``moongate'', ``pot'' for ``wooden pot'', and ``bear'' for ``teddybear''.

\myparagraph{DreamBooth~\cite{ruiz2022dreambooth}}
We use the third-party implementation~\cite{dreamboothimpl} of DreamBooth. Training is done with the frozen text transformer and fine-tuning the U-net diffusion model with a batch size of $8$ and learning rate $10^{-6}$ (without scaling with batch size and $\#$GPUs). The text prompt used for target images is {\menlo photo of [V] category} where we initialize [V] with the same rare occurring token-id $42170$ as ours. The regularization images are generated with $50$ steps of the DDPM sampler with the text prompt {\menlo photo of a \{category\}}. We train for $2500$ steps for single-concept. For multi-concept, we train for $5000$ steps but pick the best checkpoint at $3000$ iterations.
For results on CustomConcept101 dataset which we updated later in \refapp{customconcept101}, we trained DreamBooth with the learning rate of $5\times 10^{-6}$ as suggested in their paper and batch-size $4$. The training was done for $1000$ iterations for single-concept and $2000$ iterations for multi-concept.

\begin{figure*}[!t]
    \centering
    \includegraphics[width=\linewidth]{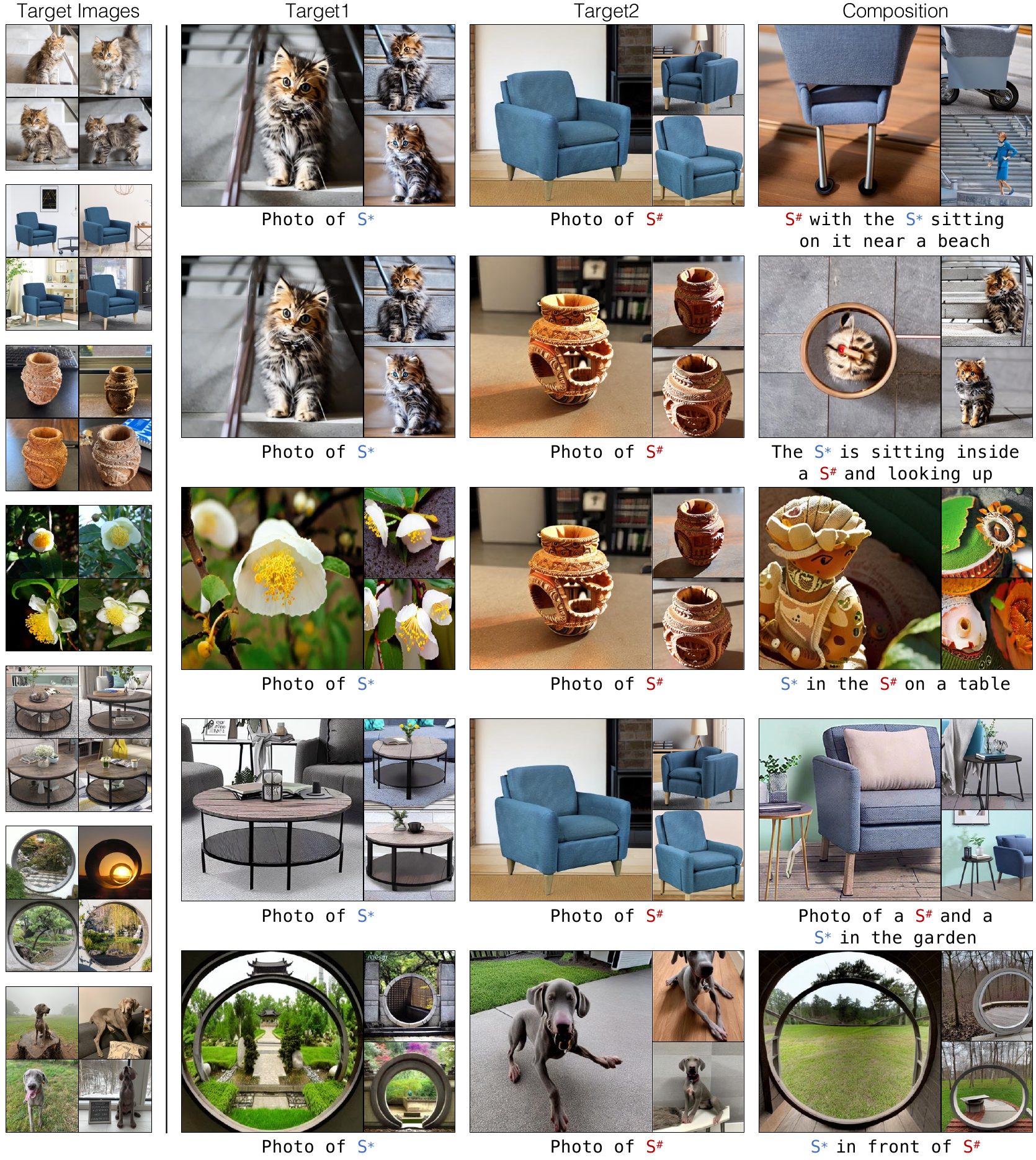}
    \vspace{-10pt}
    \caption{{\textbf{Multi-concept composition using Textual Inversion.} We observe that Textual Inversion struggles with the composition of two fine-tuned objects as shown in the above sample generations as well. 
    }}
    \label{fig:supp_ti_compose}
\end{figure*}

\begin{figure*}[!t]
    \centering
    \includegraphics[width=\linewidth]{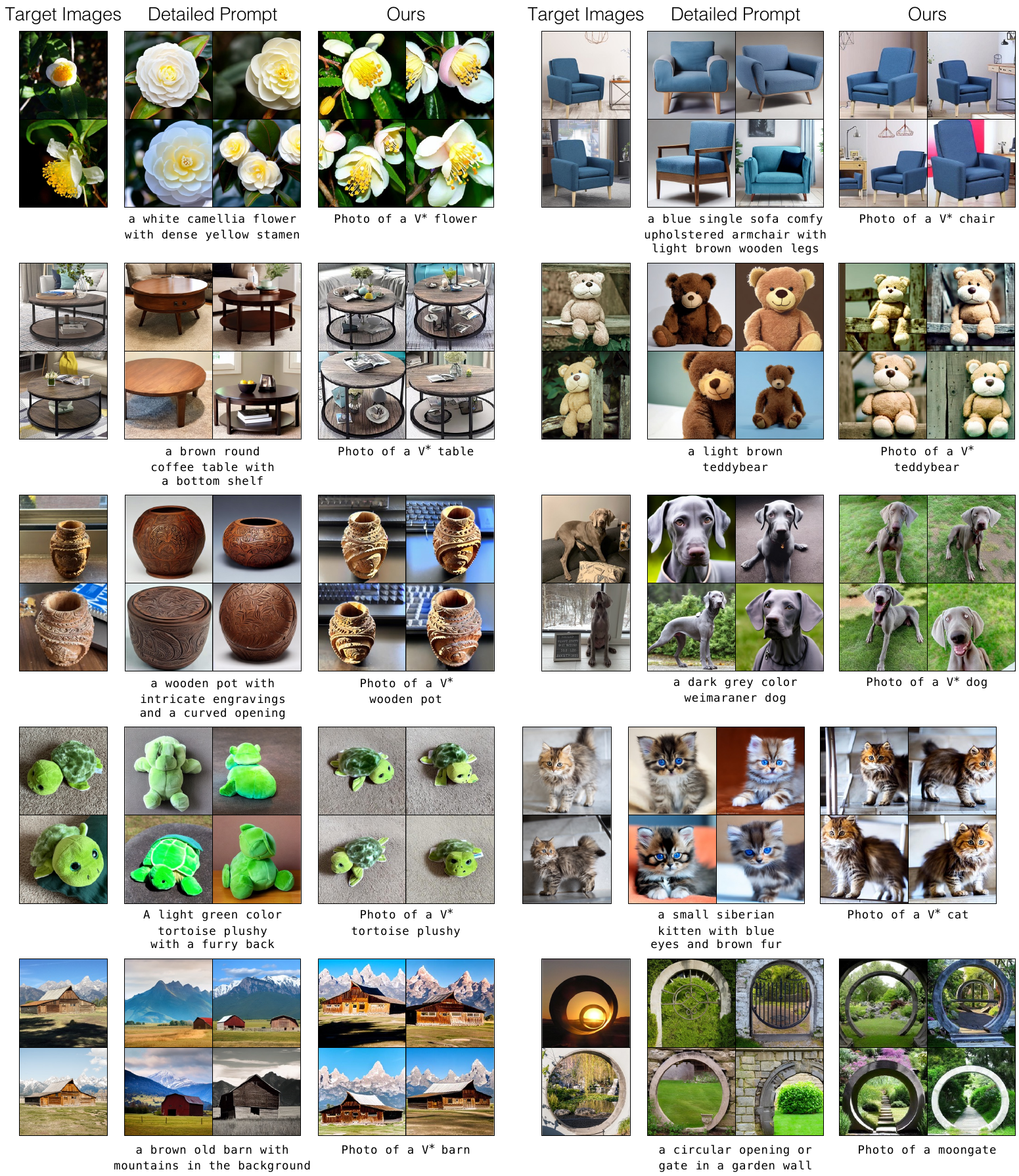}
    \vspace{-10pt}
    \caption{{\textbf{Generating target images with long text prompts in the pretrained model.} We show that even with long text descriptions the pretrained model struggles to generate exact target images. Thus, to generate the target images, we need model fine-tuning.
    }}
    \label{fig:supp_longtext}
\end{figure*}

\begin{figure*}[!t]
    \centering
    \includegraphics[width=\linewidth]{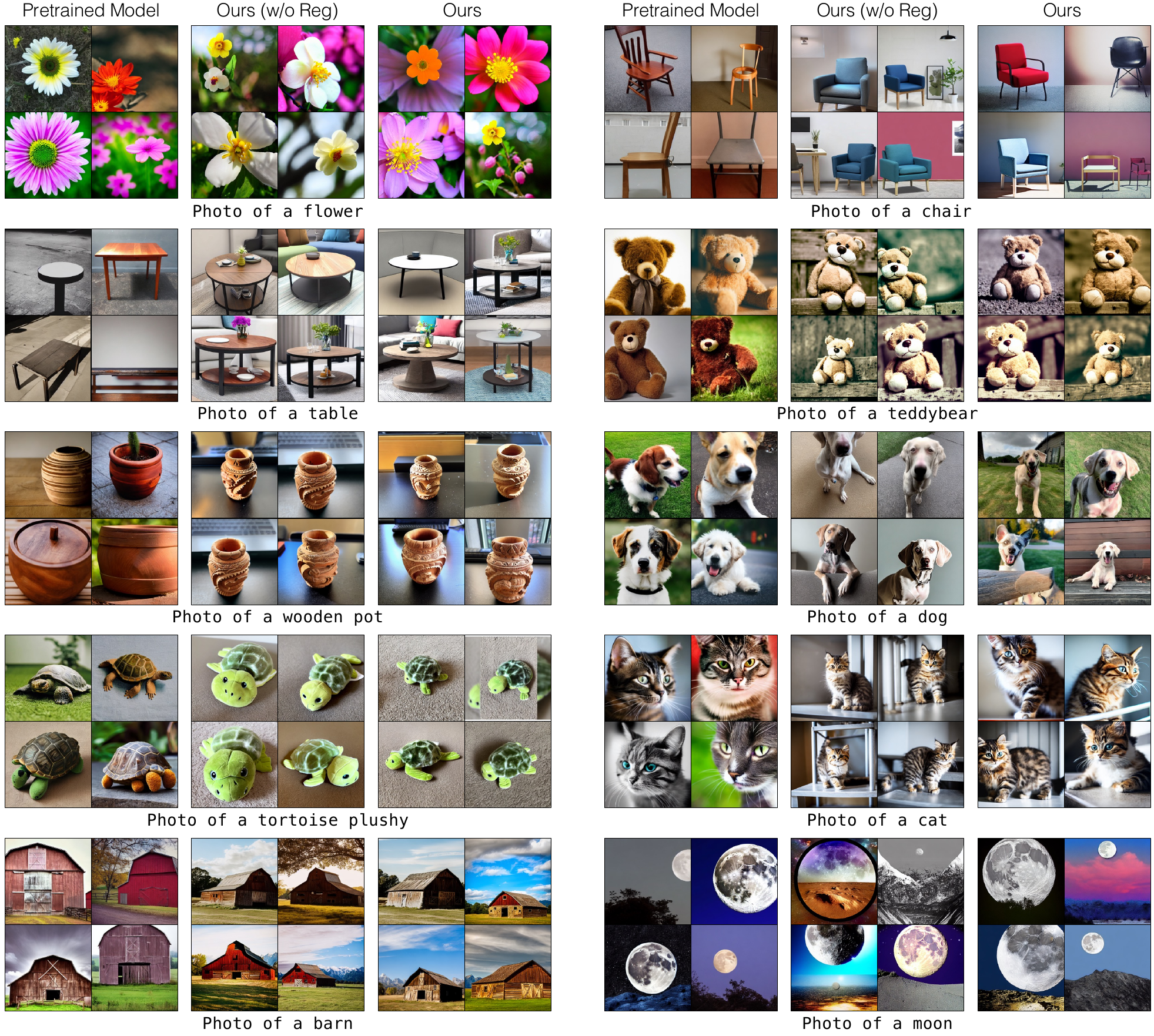}
    \vspace{-10pt}
    \caption{{\textbf{Overfitting on the training prompt template.} Since during fine-tuning, target images are trained with the text prompt, {\menlo photo of a V$^*$ \{category\}}, we show here random sample generations for the text prompt, {\menlo photo of a \{category\}}, in both ours and ours (w/o reg) case. As shown, the generations shift towards the target images and have less diversity compared to the pretrained model. Between ours and ours (w/o reg), our method has less shift and is more diverse on average. 
    }}
    \label{fig:supp_overfit}
    \vspace{-10pt}
\end{figure*}

\section{Societal Impact}\lblsec{society}
While training massive-scale diffusion models is inaccessible to most people, our method of fine-tuning pretrained models can help democratize such models to everyday users. Users can customize these models according to their own personal images, artworks, and objects of interest. Being compute and memory efficient, it will increase the accessibility and usage of large-scale models by individual users as well as enable easy collaboration for sharing millions of fine-tuned concepts and their compositions. At the same time, the dangers of generative technology accompany our method as well. Possible ways to mitigate this is the reliable detection of fake generated data, which has been studied in the context of GANs~\cite{wang2020cnn,chai2020makes} and, recently, diffusion models~\cite{corvi2022detection}.

\section{Change log}
\myparagraph{v1:} Original draft.

\myparagraph{v2:} \href{https://www.cs.cmu.edu/~custom-diffusion/dataset.html} {CustomConcept101} dataset details and results in \refapp{customconcept101}, \reffig{customconcept101}, \ref{fig:results_customconcept101}, and \reftbl{results_customconcept101}. Updated citations and results with three concept compositions in \reffig{multi_concept_limitation}.

\clearpage

\end{document}